\documentclass[journal]{IEEEtran}
\usepackage[utf8]{inputenc}   
\usepackage[T1]{fontenc}
\usepackage{textcomp}
\usepackage{amsmath,amssymb,amsfonts}

\usepackage{graphicx}
\usepackage{float}
\usepackage{multirow}
\usepackage{array}
\usepackage{hhline}
\usepackage{subfigure}
\usepackage{url}
\usepackage{multirow}
\usepackage{makecell}
\usepackage{mathrsfs}
\usepackage{amsmath,amsfonts}
\usepackage{algorithmic}
\usepackage{algorithm}
\usepackage{array}
\usepackage[caption=false,font=normalsize,labelfont=sf,textfont=sf]{subfig}
\usepackage{textcomp}
\usepackage{stfloats}
\usepackage{url}
\usepackage{verbatim}
\usepackage{graphicx}
\usepackage{cite}
\usepackage{booktabs, multirow} 
\usepackage[table,xcdraw]{xcolor}
\definecolor{table_bg}{gray}{0.93}

\usepackage{flushend}

\ifCLASSINFOpdf
\else
\fi

\begin{document}
%
\title{SceneGlue: Scene-Aware Transformer for Feature Matching without Scene-Level Annotation}
\author{Songlin~Du,
        Xiaoyong~Lu,
        Yaping~Yan,
        Guobao~Xiao,
        Xiaobo~Lu,
        and~Takeshi~Ikenaga,~\IEEEmembership{Senior~Member,~IEEE}
\thanks{This work was supported in part by the National Natural Science Foundation of China under grant 62201142, in part by the Frontier Technologies R\&D Program of Jiangsu under Grant BF2024060, in part by the Shenzhen Science and Technology Program under grant JCYJ20240813161801002, and in part by the Guangdong Basic and Applied Basic Research Foundation under grants 2025A1515011943 and 2026A1515012748. \emph{(Corresponding author: Songlin Du.)}}
\thanks{Songlin~Du, Xiaoyong~Lu and Xiaobo~Lu are with the School of Automation, Southeast University, Nanjing 210096, China, and also with the Shenzhen Research Institute of Southeast University, Shenzhen 518063, China (e-mail: sdu@seu.edu.cn; luxiaoyong@seu.edu.cn; xblu@seu.edu.cn).}
\thanks{Yaping~Yan is with the School of Computer Science and Engineering, Southeast University, Nanjing 210096, China (e-mail: yan@seu.edu.cn).}
\thanks{Guobao~Xiao is with the School of Computer Science and Technology, Tongji University, Shanghai 201804, China (e-mail: gbx@tongji.edu.cn).}
\thanks{Takeshi~Ikenaga is with the Graduate School of Information, Production and Systems, Waseda University, Kitakyushu 808-0135, Japan (e-mail: ikenaga@waseda.jp).}
}

\markboth{IEEE Transactions on Circuits and Systems for Video Technology}%
{Du \MakeLowercase{\textit{et al.}}: SceneGlue: Scene-Aware Transformer for Feature Matching without Scene-Level Annotation}

\maketitle

\begin{abstract}
Local feature matching plays a critical role in understanding the correspondence between cross-view images. However, traditional methods are constrained by the inherent local nature of feature descriptors, limiting their ability to capture non-local scene information that is essential for accurate cross-view correspondence. In this paper, we introduce SceneGlue, a scene-aware feature matching framework designed to overcome these limitations. SceneGlue leverages a hybridizable matching paradigm that integrates implicit parallel attention and explicit cross-view visibility estimation. The parallel attention mechanism simultaneously exchanges information among local descriptors within and across images, enhancing the scene's global context. To further enrich the scene awareness, we propose the Visibility Transformer, which explicitly categorizes features into visible and invisible regions, providing an understanding of cross-view scene visibility. By combining explicit and implicit scene-level awareness, SceneGlue effectively compensates for the local descriptor constraints. Notably, SceneGlue is trained using only local feature matches, without requiring scene-level groundtruth annotations. This scene-aware approach not only improves accuracy and robustness but also enhances interpretability compared to traditional methods. Extensive experiments on applications such as homography estimation, pose estimation, image matching, and visual localization validate SceneGlue's superior performance. The source code is available at https://github.com/songlin-du/SceneGlue.
\end{abstract}


\IEEEpeerreviewmaketitle

\section{Introduction}
\IEEEPARstart{F}{eature} matching, which aims to find the correct correspondence between two sets of features, is a fundamental problem for many computer vision tasks, such as object recognition \cite{Survey2025}, structure from motion (SfM) \cite{10148998}, and simultaneous localization and mapping (SLAM) \cite{10742393}. Existing feature matching approaches mainly resort to local descriptors to represent textures around the keypoints of neighboring regions of interest \cite{sift,d2net,superpoint}. Specifically, a general pipeline consists of four phases: 1) feature detection, 2) feature description, 3) feature matching, and 4) outlier filtering. In the feature detection part, keypoints that have distinguishable features to facilitate matching are detected. For feature description, descriptors are extracted based on keypoints and their neighborhoods. The keypoint positions and the corresponding descriptors are employed as features of the image. Then the feature-matching algorithm is applied to find the correct correspondence in two sets of extracted features. Finally, the outlier filtering algorithm is applied to identify and reject outlier matches based on the obtained matches \cite{MatchMamba,CGR-Net,MGCNet,U-Match}.

\begin{figure}[t]
	\includegraphics[width=0.99\linewidth]{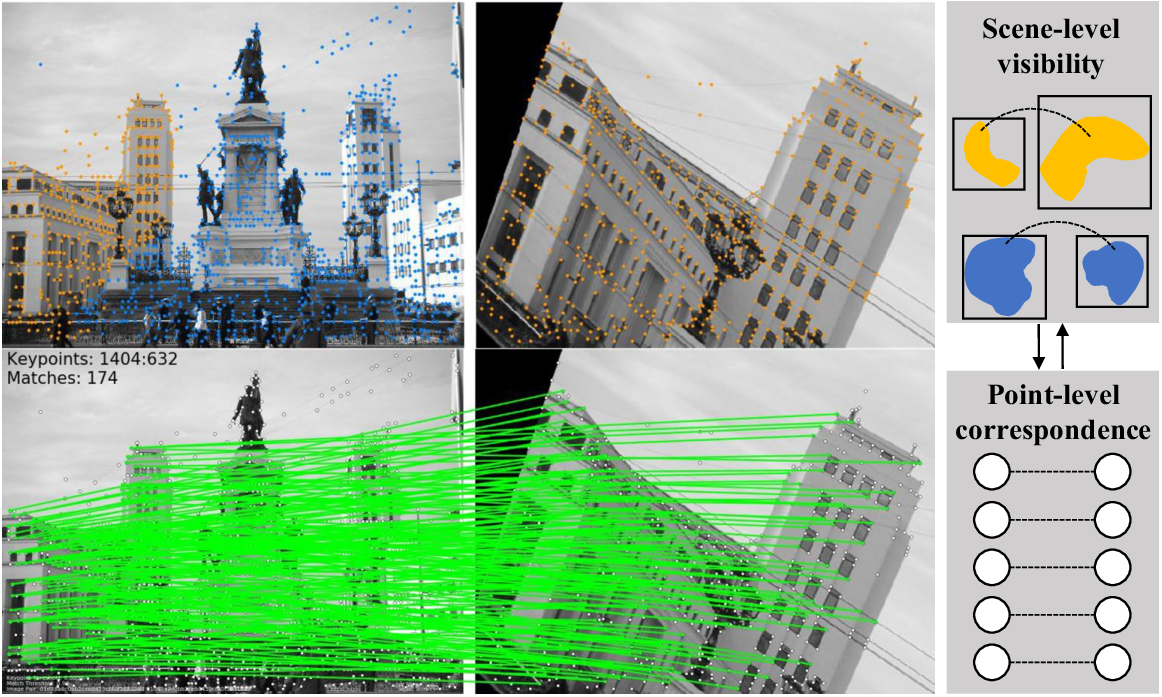}
	\centering
	 \caption{Graphical illustration of the intuition lying behind the proposed SceneGlue. The scene-level estimation of cross-view visibility would enable the local feature descriptors to overcome the limitation caused by their local nature.}
	 \label{figure1}
\end{figure}

Due to the local nature of feature descriptors, it is challenging to find invariance and obtain robust local correspondences between cross-view images of large-scale scenes \cite{MINIMA}. Benefitting from the rapid advancement of deep learning, the state-of-the-art feature matching approaches learn feature descriptors by convolutional neural networks (CNN) to achieve better representation capability \cite{superpoint,ASLFeat} rather than design handcrafted feature representations. Dominant methods infer the correspondence relations among learned cross-view features by graph neural networks and attention mechanisms \cite{superglue,loftr,LightGlue}, because attention-based neural networks are able to enhance local features by perceiving global information and modeling contextual relationships based on local features.
Recalling the behavior of humans making correspondence between cross-view images, we naturally follow the common manner of observing from global scene-level semantics to local point-level pixels. It is clear that the matching pipeline of existing methods conflicts with human style. Therefore, this work aims to extract scene-aware correspondence information and integrate it with local feature correspondence in an end-to-end learning framework that is more accordant to the biological habit from billions of years of evolution and is expected to achieve superior performance over existing approaches.

Driven by above motivations, this paper proposes SceneGlue that innovates throughout the feature-matching framework. In particular, SceneGlue makes the following contributions:
\begin{itemize}
  \item SceneGlue builds local feature descriptors from an informative feature representation strategy that introduces a wave position encoder to dynamically adjusts the relationship between descriptor and position by amplitude and phase. Compared to existing position encoding methods, the position is represented as a wave and is unfolded into real parts and imaginary parts to efficiently encode position information.
  \item SceneGlue arranges self- and cross-attentions in parallel to implicitly understand scene-level correspondence. Compared to sequential self- and cross-attention, the parallel self- and cross-attention enables more sufficient interactions of features from inter- and intra-images. SceneGlue therefore achieves higher matching accuracy and higher efficiency simultaneously.
  \item SceneGlue proposes a Visibility Transformer to predict cross-view visible regions with a classification manner, rather than selecting image tokens as group tokens to represent scene-level correspondence like existing works. The Visibility Transformer is more explicable and natural in understanding scene-level overlapping regions in cross-view images without additional scene-level groundtruth.
\end{itemize}
These contributions make SceneGlue to achieve superior performance over existing feature matching methods on various applications, including image matching, homography estimation, outdoor pose estimation, and outdoor visual localization.

The rest of this paper is organized as follows. Section \ref{relateworks} reviews related works. The proposed SceneGlue is presented in Section \ref{sec:Methodology}. Experimental results and analysis are presented in Section \ref{experiments}. Limitations of the proposed SceneGlue and future works are discussed in Section \ref{Limitation}. Finally, the conclusion is drawn in Section \ref{conclusion}.

\section{Related Works}
\label{relateworks}
\subsection{Local Feature Matching}
\noindent{\textbf{Robust Feature Representation.}} Early local feature matching methods mainly rely on handcrafted keypoint detectors and descriptors \cite{sift,orb}, which achieved remarkable success and have been widely applied in many 3D vision tasks. With the development of deep learning, learning-based detectors and descriptors have been proposed to improve robustness under illumination and viewpoint changes \cite{r2d2,superpoint,d2net}. Benefiting from deep networks and large-scale data, learned features adapt well to diverse textures, lighting conditions, and viewpoints, often outperforming handcrafted ones. Despite these advances, two challenges remain: 1) Learning scale invariance from large-scale data to address scale inconsistency across views; and 2) Effectively encoding keypoint positions into descriptors so that geometric structure can be better exploited during matching.

\noindent{\textbf{Position Encoding.}}
Position encoding aims to incorporate spatial information into feature representations so that descriptors become aware of relative or absolute positions. Early approaches \cite{transformer} use fixed sinusoidal functions or learnable vectors added to the original features. While simple, these methods restrict model flexibility because the encoding length is fixed during training, limiting input size at inference. Relative position encoding \cite{swin} adjusts attention weights based on relative positions, but it is computationally expensive and requires interpolation for varying input lengths, which may degrade performance. Convolution-based encoders \cite{matchformer} introduce positional cues through convolution, but they are limited to Euclidean data such as feature maps and cannot handle sparse descriptors. To support arbitrary-length and non-Euclidean inputs, SuperGlue adopts an MLP-based encoder that maps coordinate vectors to the descriptor dimension. However, its encoding capacity remains limited. Inspired by Wave-MLP \cite{wave-mlp}, we treat descriptors as amplitude and positions as phase, and fuse them via the Euler formula to produce position-aware descriptors.

\noindent{\textbf{Matching.}}
Recent feature matchers learn correspondences between two sets of local features using attention-based Graph Neural Networks (GNNs) \cite{10794561,oetr}.
SuperGlue \cite{superglue} is a pioneering work that introduces an attention-based matching network, where self-attention aggregates contextual information within each image and cross-attention identifies potential correspondences across images.
LightGlue \cite{LightGlue} extends this framework by incorporating confidence estimation and adaptive computation. Through early-exit layers and pruning of unlikely matches, it improves speed and training efficiency while maintaining accuracy as a drop-in replacement for SuperGlue. Compared with LightGlue, the proposed SceneGlue emphasizes scene-aware matching by integrating global visibility estimation and parallel attention, improving robustness in challenging conditions. While LightGlue focuses on computational efficiency through adaptive depth and pruning, SceneGlue explicitly models scene-level relationships.

Beyond sparse descriptors, LoFTR \cite{loftr} applies self- and cross-attention directly on CNN feature maps and produces matches in a coarse-to-fine manner. To estimate dense correspondences, Truong \emph{et al.} \cite{Truong2021} model the predictive distribution with a constrained mixture formulation to better capture accurate flow and outliers. MatchFormer \cite{matchformer} further removes the CNN backbone and adopts a pure attention architecture that jointly performs feature extraction and matching, with interleaving attention that emphasizes self-attention early and cross-attention later. DeepMatcher \cite{DeepMatcher} employs a lightweight Transformer to model long-range keypoint relationships, while VD-Matcher \cite{VD-Matcher} introduces weight recycling and a lightweight multi-scale keypoint detection module to control model size and computation. ASTR \cite{ASTR} proposes an adaptive spot-guided Transformer that addresses local consistency and scale variation through a unified coarse-to-fine architecture with spot-guided aggregation and depth-aware scaling.

Most of these methods model relationships at the point level and do not explicitly leverage higher-level scene information, which can lead to degraded performance in challenging scenarios such as wide-baseline images \cite{10794561}. To address this limitation, we introduce a Visibility Transformer that predicts cross-view visible regions and enables scene-level correspondence learning without additional scene annotations. In contrast to previous methods where self- and cross-attention are applied sequentially, our framework performs them in parallel.

\subsection{Scene Awareness}
Since local feature descriptors are extracted from a local block in the image, they are naturally lacking the perception ability on non-local scenes. \emph{Scene awareness} is essential for matching local features extracted from cross-view images. However, this problem has not attracted widespread attention in this field. Existing approaches mainly try to improve the quality of feature descriptors for a more comprehensive representation of both local information and non-local context information. As a representative method, ContextDesc \cite{ContextDesc} achieves cross-modality context representation by a unified learning framework that leverages and aggregates visual context from high-level image representation and geometric context from keypoint distribution. Some methods also try to enhance the context representation capability by fusing features of more modalities or more levels. ASLFeat \cite{ASLFeat} enables conventional local feature descriptors to be aware of the shape and achieves stronger geometric invariance by fusing shape features.
Context Augmentation \cite{Changwei23} employs the visual Transformer and a learnable gated map to adaptively embed global context and location information into local descriptors. Different from the above approaches that extend existing feature descriptors from various aspects, state-of-the-art feature matchers also pay attention to the visibility of cross views. OETR \cite{oetr} applies a DETR-like model to detect the common view area of two images and scales the common view area to match at the same scale. OAMatcher \cite{OAMatcher} prioritizes overlapping regions in image pairs during descriptor enhancement, ensuring minimal disturbance from non-overlapping areas. AdaMatcher \cite{AdaMatcher} overcomes the limitations of mutual nearest neighbor by utilizing an elaborate feature interaction module for feature correlation and co-visible area estimation. It performs adaptive assignment on patch-level matching, estimates image scales, and refines co-visible matches through scale alignment and sub-pixel regression. SAM \cite{XiaoyongICCV2023} takes point-level descriptors as image tokens and introduces the concept of group tokens that represent a group of features shared by cross-view images. Although Transformer does have the ability to understand the relationship between local feature descriptors and scene-level features, straightforwardly grouping tokens is not the optimal way of understanding scene-level correspondence for local feature matching.

Although existing researches studied both feature descriptors and matchers to incorporate the context of local features or implicitly exploit the relations among keypoints for a more intelligent perception of global scene information \cite{li2025comatch}, a more explicit framework that bridges the gap between local features and the global scene is still highly desired. This paper highlights the significance of scene awareness in local feature matching, as most existing methods primarily enhance keypoints using deep neural networks.
In particular, the proposed SceneGlue explicitly models scene awareness through learnable scene descriptors, which indicate scene-level visibility across the entire image, rather than being limited to keypoint-level local features.

\section{Methodology}
\label{sec:Methodology}
Assume that two sets of features $d_s\in\mathbb{R}^{M\times C}$, $d_t\in\mathbb{R}^{N\times C}$ need to be matched which are descriptors extracted from two images.
Subscripts $s$ and $t$ stand for the source image and target image, respectively.
$M$, $N$ are the number of descriptors, $C$ is the channels of descriptors.
The keypoint positions are denoted as $p_s\in\mathbb{R}^{M\times 3}$, $p_t\in\mathbb{R}^{N\times 3}$. For each position $p=[p_u, p_v, p_c]$, the $[p_u, p_v]$ denotes coordinates of the keypoint in image, and $p_c$ denotes the confidence value of the keypoint detector. Our objective is to find the correct correspondence between the two images utilizing descriptors and position information.

An overview of the network architecture is shown in Fig. \ref{arch:figure}. SceneGlue defines both learnable local descriptors and scene descriptors for each input image. Each multi-scale local descriptor is obtained by embedding the position of the corresponding keypoint into local image appearance features learned by an off-the-shelf descriptor through the proposed wave position encoding. The position-aware multi-scale local descriptors are concatenated with the learnable scene descriptors to perform parallel attention. The learnable scene descriptor is a set of learnable parameters with size ${2\times C}$ in the proposed framework. In training phase, the learnable scene descriptors first randomly initialized in the same way with other learnable parameters; In testing phase, they are fed to the proposed Visibility Transformer for cross-view visible area estimation. In the Parallel Attention module, the self-attention layer and cross-attention layer are applied for $L$ times to utilize the global context to enhance the features, which are then re-split to go through the proposed Visibility Transformer that is able to estimate scene-level correspondence. The overall framework is supervised by keypoint-level groundtruth only and does not require scene-level annotation.

\subsection{Informative Feature Representation}
\subsubsection{Multi-scale Local Descriptor}
To enhance the robustness of our matcher in scale-varying scenarios, we design an additional lightweight multi-scale feature network based on the SuperPoint detector \cite{superpoint}.
As shown in Fig. \ref{figure3}, in addition to keypoint sampling on $1/8$ resolution feature maps as in SuperPoint, we also perform keypoint sampling on $1, 1/2, 1/4$ resolution feature maps and fuse features on 4 scales as descriptors using lightweight linear layers. With the introduction of a small number of parameters and computations, the multi-scale feature network improves the performance of the matcher significantly.

\begin{figure*}[t]
	\includegraphics[width=0.99\linewidth]{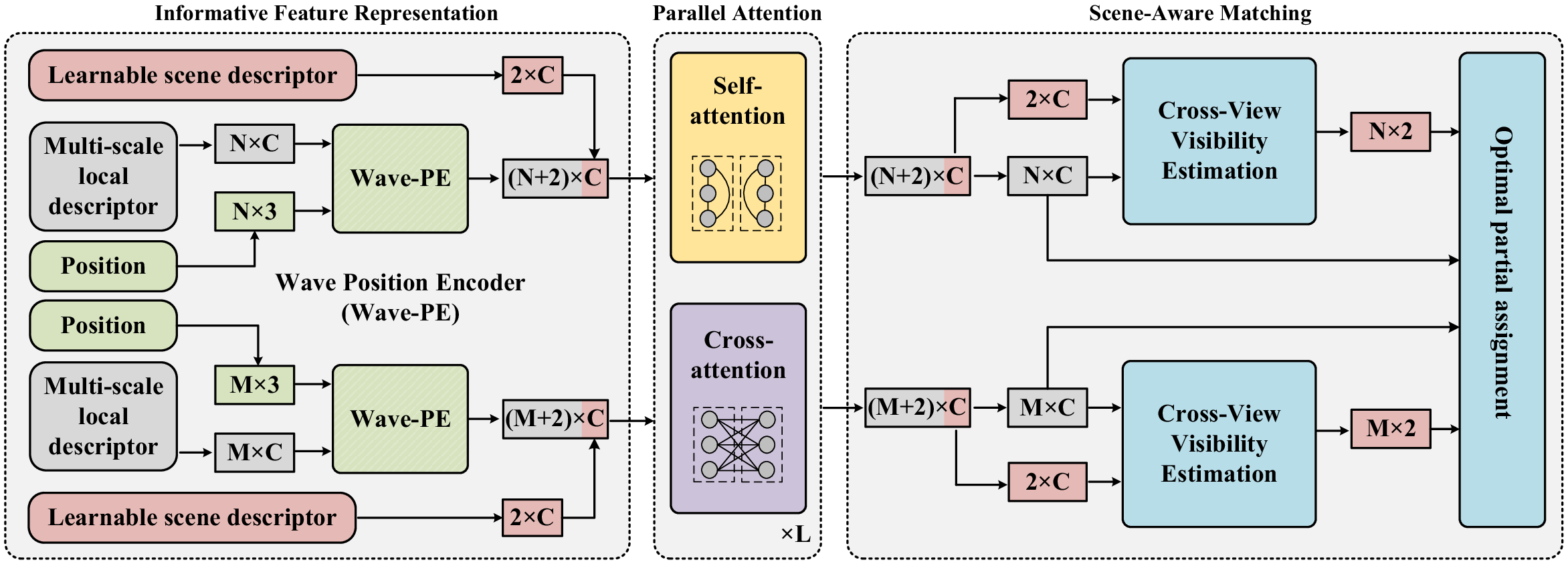}
	\centering
	 \caption{
	 \textbf{Graphical illustration of the proposed SceneGlue.} The proposed scene-aware matching method consists of three parts, namely informative feature representation, parallel attention, and scene-aware matching. The informative feature representation first encodes the position by a Wave Position Encoder (Wave-PE) to obtain position-aware descriptors and then combines each local position-aware descriptor with a learnable scene descriptor. The feature descriptors are passed through the proposed parallel attention layers to implicitly achieve scene awareness. Finally, the leveraged features are utilized for scene-level cross-view visibility estimation and point-level feature matching simultaneously.}
	 \label{arch:figure}
\end{figure*}

\subsubsection{Wave Position Encoder}
\label{Wave position encoder}
For the MLP-based position encoder (MLP-PE), the main drawback is the limited encoding capacity because
the parameters of MLP-PE are less than 1$\%$ of the whole network, yet the position information is important for feature matching.
Therefore, a Wave Position Encoder (Wave-PE) is designed to dynamically adjust the relationship between descriptor and position by amplitude and phase to obtain better position encoding.

In Wave-PE, position encoding is represented as a wave $\tilde{{w}}$ with both amplitude ${A}$ and phase ${\theta}$ information,
and the Euler formula is employed to unfold waves into real parts and imaginary parts to process waves efficiently,
\begin{equation}
\begin{split}
	\tilde{{w}}_{j} & ={A}_{j} \odot e^{i{\theta} _{j} } \\
	 & ={A}_{j} \odot \cos {\theta}_{j}+i{A}_{j} \odot \sin {\theta}_{j} ,j=1,2,...,n.
	\label{equ1}
\end{split}
\end{equation}
As shown in Fig. \ref{ParaAttention}, the amplitude and phase are estimated by two learnable networks via descriptors and position, respectively.
Then a learnable network is applied to fuse the real and imaginary parts into position encoding,
\begin{equation}
\begin{split}
	&{A}_{j}= \mathrm{MLP_{A}}({d} _{j}), \\
	&{\theta} _{j}= \mathrm{MLP_{\theta}}({p}_{j}), \\
	&{x}_{j}^{0}= {d}_{j}+\mathrm{MLP_{F}}([{A}_{j} \odot \cos {\theta}_{j},{A}_{j} \odot \sin {\theta}_{j}]).
	\label{equ:waveposition}
\end{split}
\end{equation}
\noindent$[\cdot,\cdot]$ denotes concatenation.
For three learnable networks in equation (\ref{equ:waveposition}), two-layer MLP is chosen for simplicity.

\subsection{Parallel Attention for Implicit Scene-Awareness}
\label{sec:Parallelattention}
Proper integration of self-attention and cross-attention leverages the global learning ability and good parallelism, which is able to further highlight the critical information in the implicit scene-aware representations while suppressing the useless noise. Therefore, we propose to integrate self-attention and cross-attention in a parallel manner, which is the so-called \emph{parallel attention} in this paper.

As illustrated in the right part of Fig. \ref{ParaAttention}, to utilize self-attention and cross-attention to enhance the scene-level representation ability of the multi-scale local feature descriptors, the two sets of descriptors from two images are first linearly projected to query, key and value, \emph{i.e.}, (${Q}_{s}, {K}_{s}, {V}_{s}$) and (${Q}_{t}, {K}_{t}, {V}_{t}$), respectively.
Then self- and cross-attention are computed in a parallel manner.
In the self-attention module, the standard attention computation $\mathrm{softmax}({Q}{K}^{\top}/\sqrt{d}){V}$ is employed,
where ${Q}, {K}, {V}$ come from the same input, \emph{i.e.},
(${Q}_{s}, {K}_{s}, {V}_{s}$) or (${Q}_{t}, {K}_{t}, {V}_{t}$).
In the cross-attention module, an attention weight-sharing strategy is proposed to improve model efficiency. Existing cross-attention approaches follow
\begin{equation}
\begin{cases}
s\to t:\; \mathrm{softmax}({Q}_{t}{K}_{s}^{\top}/\sqrt{d}){V}_{s}\\
t\to s:\; \mathrm{softmax}({Q}_{s}{K}_{t}^{\top}/\sqrt{d}){V}_{t},
\end{cases}
\end{equation}
which are computationally costly, since the cross-attention has to be calculated twice. Unlike existing cross-attention, the proposed parallel attention suggests a novel paradigm
\begin{equation}
\begin{cases}
s\to t:\; \mathrm{softmax}(({Q}_{s}{K}_{t}^{\top})^{\top}/\sqrt{d}){V}_{s}\\
t\to s:\; \mathrm{softmax}(({Q}_{s}{K}_{t}^{\top})/\sqrt{d}){V}_{t},
\end{cases}
\end{equation}
in which the input is (${Q}_{s}, {V}_{s}, {K}_{t}, {V}_{t}$). This paradigm is more efficient since it does not compute round-trip cross attention.

\begin{figure}[t]
	\includegraphics[width=0.8\linewidth]{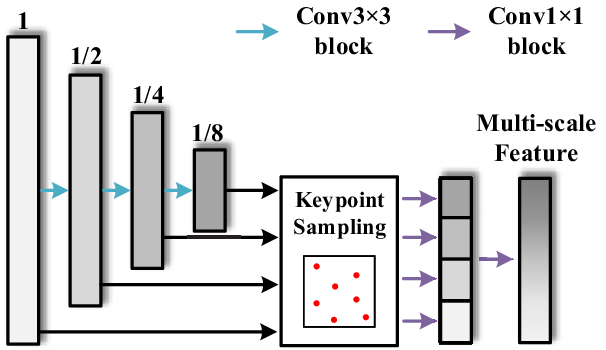}
	\centering
		\caption{\textbf{Multi-scale Feature Detector.} Features in $1, 1/2, 1/4, 1/8$ resolution are sampled and fused by lightweight linear layers.
		The Multi-scale feature network allows feature matching to be more robust to large-scale variation scenarios.
		}
		\label{figure3}
\end{figure}

\begin{figure*}[t]
    \centering
    \includegraphics[width=0.8\textwidth]{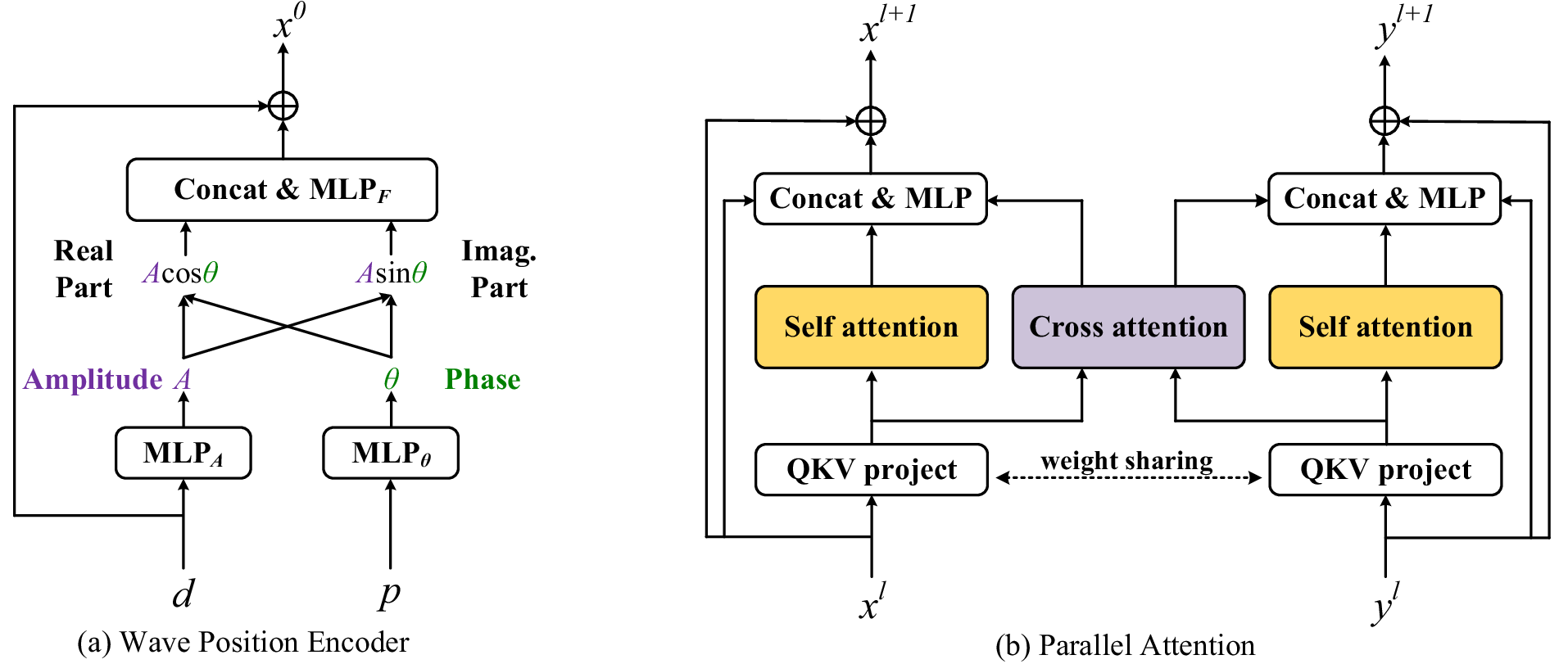} 
    \caption{\textbf{Graphical illustration on the wave position encoder and the parallel attention}.
	(a) The wave position encoder fuses the amplitude ${A}$ estimated with the descriptor ${d}$ and the phase ${\theta}$ estimated with the position ${p}$ to generate position encoding.
	(b) Stacked parallel attention layers utilize self- and cross-attention to enhance the descriptors and find potential matches, where self- and cross-attention are adaptively integrated through a learnable network.}
    \label{ParaAttention}
\end{figure*}

Different from sequential manner and alternating manner, the self-attention and cross-attention are arranged in a parallel manner, and fused by a two-layer MLP in this work. The unique parallel framework makes three key contributions compared with existing works:
\begin{enumerate}
  \item Parallel attention enables local feature descriptors of intra- and inter-images to interact with each other simultaneously. The simultaneous interaction mines more critical information in the implicit scene-aware representations which are crucial to final performance.
  \item Parallel attention saves redundant parameters and computations while boosting performance through a learnable fusion.
  \item Parallel attention can be used as a pluggable module since it updates two sets of descriptors simultaneously. We can simply stack \emph{L} parallel attention layers to enhance representation ability
and conveniently explore various architectures to design model variants.
\end{enumerate}

\subsection{Cross-View Visibility Estimation for Explicit Scene-Awareness}
\label{sec:Visibility}
\subsubsection{Cross-View Visibility Estimation}
\begin{figure}[tb]
	\includegraphics[width=0.7\linewidth]{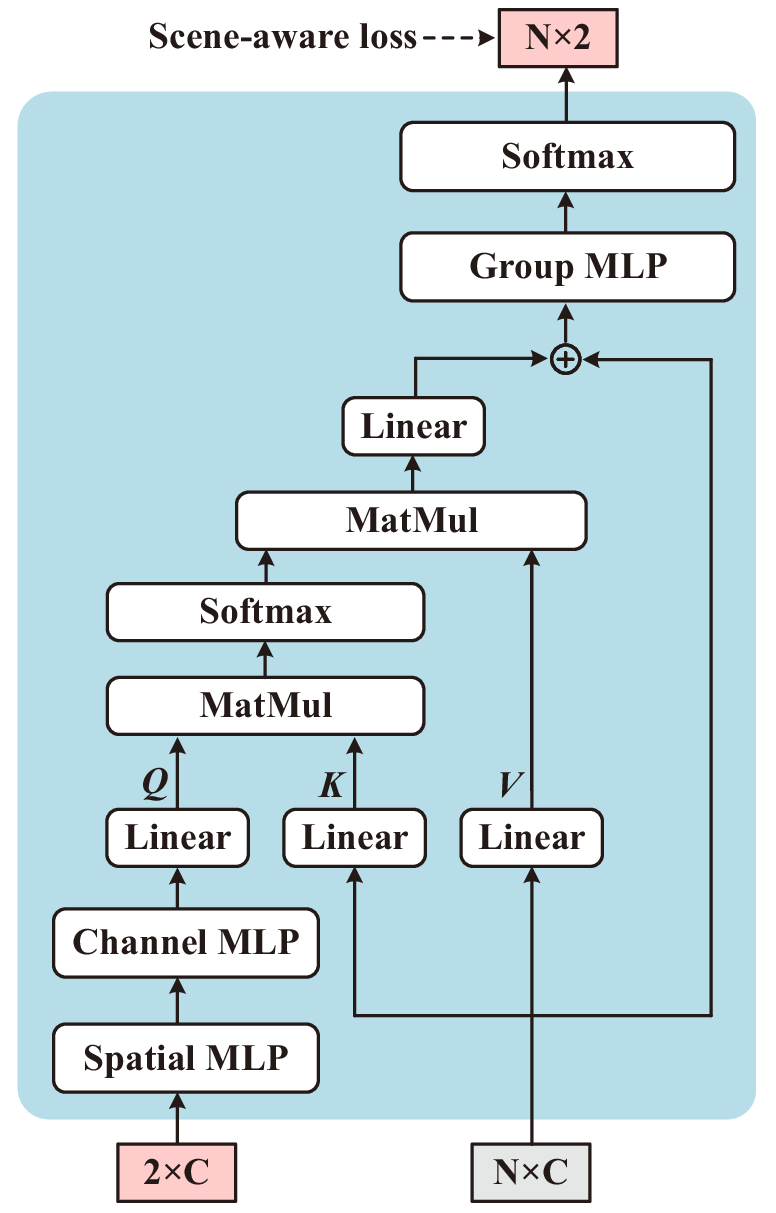}
	\centering
		\caption{\textbf{Visibility Transformer.} The Visibility Transformer is proposed for cross-view visible area estimation. It adopts a Transformer architecture to establish the relationship between multi-scale local descriptors and learnable scene descriptors before matching and assigns the multi-scale local descriptors to the commonly-visible area or commonly-invisible area through a $\mathrm{Softmax}$ classifier.}
		\label{figure4}
\end{figure}

As shown in Fig. \ref{figure4}, to estimate the commonly visible area in images taken from different views, a variation of Transformer, entitled Visibility Transformer, is proposed in this section. The commonly-visible area estimation method takes the multi-scale local descriptor and learnable scene descriptor as input, and employs three modules, namely spatial MLP, channel MLP, and the proposed Visibility Transformer, to predict the values of the final learnable scene descriptor.

Spatial MLP and channel MLP are introduced to enhance the capacity of the cross-view visibility estimation module. Denoted by $W_{1}^{s}, W_{1}^{c},W_{2}^{s}, W_{2}^{c}$ the learnable weight matrices, and $\sigma$ is the Gaussian Error Linear Unit (GELU) function, the spatial MLP is defined by
\begin{equation}
O_{*,i}=I_{*,i}+W_{2}^{s}\sigma(W_{1}^{s}I_{*,i}),
\end{equation}
which transposes learnable scene descriptors and maps $\mathbb{R}_{2} \mapsto \mathbb{R}_{2}$ to enable interaction between the learnable scene descriptors. The channel MLP is defined by
\begin{equation}
O_{j,*}=I_{j,*}+W_{2}^{c}\sigma(W_{1}^{c}I_{j,*}),
\end{equation}
which maps $\mathbb{R}_{C} \mapsto \mathbb{R}_{C}$ to interact between channels.

Unlike the standard Transformer that projects the same input to ${Q}, {K}, {V}$, the proposed Visibility Transformer establishes the relationship between informative local descriptors and learnable scene descriptors by linearly projecting each learnable scene descriptor as query matrix ${Q}$, and each local descriptor as key matrix ${K}$ and value matrix ${V}$. The three matrices derived from different inputs are then packed together by
\begin{equation}
b_{k} = \mathrm{softmax}({Q}{K}^{\top}/\sqrt{d}){V}.
\end{equation}
The $b_{k}$ is then added to the original informative descriptor after a linear projection. The fused features are projected $\mathbb{R}_{N \times C} \mapsto \mathbb{R}_{N \times 2}$ by a group MLP to make preparation for learning the cross-view visible regions.

In physical scenes, the cross-view visible regions and non-visible regions have clear boundaries in images. We therefore model the task of recognizing cross-view visible regions as a classification problem. Based on this motivation, a simple yet effective way is to make the last $\mathrm{Softmax}$ layer learns the values of the learnable scene descriptor through a binary cross-entropy loss. In particular, the input informative local descriptors are classified as commonly visible ones and invisible ones through comparing the learned scene descriptors with groundtruth by
\begin{equation}
  \mathcal{L}_{scene} = \frac{1}{N}\sum_{k}-\left[ y_{k} \log b_{k} + \left( 1-y_{k} \right) \log \left( 1-b_{k} \right) \right] ,
  \label{equ12}
\end{equation}
where $N$ is the number of descriptors, $b_{k}$ is the $k$th learned floating-point vector indicating whether the $k$th keypoint is visible cross views or not, and $y_{k}$ is the corresponding groundtruth of $b_{k}$. The Visibility Transformer forms a scene-aware supervision signal as it distinguishes the visible/non-visible regions in cross-view scenes, and therefore encourages the corresponding keypoint in the visible regions to match with each other and avoids keypoint in the non-visible regions from being matched. With the hybrid loss function supervised by groundtruth matches, we successfully train the learnable scene descriptor to guide the matching without relying on scene-level supervision.

\subsubsection{Optimal Partial Assignment}
As illustrated in Fig. \ref{arch:figure}, the optimal partial assignment takes two kinds of features from each view, \emph{i.e.}, the multi-scale local descriptors leveraged by parallel attention and the learned scene descriptors. Specifically, the multi-scale local descriptors leveraged by parallel attention are for local feature matching; the learned scene descriptors are for calculating the scene-aware loss term $\mathcal{L}_{scene}$ defined in (\ref{equ12}). The feature-matching task is achieved by finding an optimal partial assignment between two sets of features. To this end, a point-level score matrix $\mathbf{S}\in\mathbb{R}^{M\times N}$ is computed from the two sets of learned descriptors, \emph{i.e.},
\begin{equation}
	\begin{split}
		&\mathbf{S}_{i,j}=<f^{s}_{i},f^{t}_{j}>,
		\label{equ9}
	\end{split}
\end{equation}
where $<\cdot,\cdot>$ denotes inner product. As opposed to learned visual descriptors, the matching descriptors are not normalized, and their magnitude can change per feature and during training to reflect the prediction confidence.

For supervision, groundtruth set of matched keypoints \emph{$\mathcal{M} _{gt}$}, groundtruth set of unmatched keypoints in source image \emph{$\mathcal{D}_{s}$} and groundtruth set of unmatched keypoints in target image \emph{$\mathcal{D}_{t}$} are computed from groundtruth homography or groundtruth camera pose. The parallel attention layers and the Visibility Transformer are optimized by the negative log-likelihood loss over the optimal partial assignment matrix $\mathbf{P}$, which is denoted as
\begin{equation}
\begin{split}
  \mathcal{L}_{feature} = &-\sum_{(i,j)\in \mathcal{M}_{gt}} \log \bar{\mathbf{P}}_{i,j} - \sum_{i\in \mathcal{D}_{s}} \log \bar{\mathbf{P}}_{i,N+1}  \\
  &- \sum_{j\in \mathcal{D}_{t}} \log \bar{\mathbf{P}}_{M+1,j}.
\end{split}
  \label{equ11}
\end{equation}

Unlike existing method only computing the dot product of two sets of local feature descriptors as a score matrix that ignores the scene awareness, the proposed method takes both descriptor similarity and scene awareness into consideration, resulting in a hybrid loss function $\mathcal{L}=\mathcal{L}_{feature}+\alpha\mathcal{L}_{scene}$ which is composed of the descriptor-similarity term $\mathcal{L}_{feature}$ defined in (\ref{equ11}) and the scene-aware term $\mathcal{L}_{scene}$ defined in (\ref{equ12}).

\subsection{Discussions}
SAM enhances feature matching by introducing group tokens, which represent shared features across images and are modeled using a Transformer to assign corresponding points to the same group while distinguishing non-corresponding points. SceneGlue builds upon SAM \cite{XiaoyongICCV2023} by fundamentally improving the modeling of scene-level correspondences for feature matching. While SAM introduces group tokens and a Transformer-based grouping mechanism to construct scene-aware features, SceneGlue enhances this framework through three key innovations: (1) a multi-resolution informative feature representation with a wave position encoder to improve scale invariance; (2) a parallel self- and cross-attention structure to implicitly capture scene-level relationships with higher accuracy and efficiency; and (3) a visibility Transformer that directly predicts overlapping regions across views, offering a more interpretable alternative to token grouping. These improvements are essentially different from existing scene-aware attempt \cite{oetr,OAMatcher,AdaMatcher,XiaoyongICCV2023}, enabling SceneGlue to achieve superior performance across multiple downstream tasks.

\begin{table*}[htb]
	\centering
    \caption{\textbf{Image matching performance on the HPatches dataset.}  The \textbf{best} results are highlighted in bold.}
	\begin{tabular}{clcccccccccc}
	\toprule
		&\multicolumn{1}{c}{\multirow{2}{*}{Matcher}} &\multicolumn{10}{c}{Mean Matching Accuracy} \\
		&  &{@1px} &{@2px} &{@3px} &{@4px}  &{@5px} &{@6px}  &{@7px} &{@8px} &{@9px} &{@10px} \\
		\midrule
        \multirow{11}{*}{Viewpoint change}
        &SIFT+CAPS &0.32 &0.50 &0.60 &0.68 &0.74 &0.78 &0.80 &0.82 &0.83 &0.84\\
        &D2Net &0.06 &0.19 &0.34 &0.47 &0.57 &0.64 &0.69 &0.72 &0.74 &0.76\\
        &SuperPoint &0.25 &0.49 &0.61 &0.67 &0.70 &0.72 &0.73 &0.74 &0.75 &0.75\\
        &SuperPoint+CAPS &0.27 &0.54 &0.68 &0.74 &0.78 &0.80 &0.82 &0.83 &0.84 &0.85\\
        &R2D2 &0.29 &0.59 &0.71 &0.76 &0.79 &0.80 &0.81 &0.81 &0.82 &0.82 \\
        &Patch2Pix &0.29 &0.59 &0.73 &0.80 &0.84 &0.86 &0.87 &0.88 &0.89 &0.90 \\
        &NCNet &0.02 &0.08 &0.17 &0.27 &0.38 &0.47 &0.55 &0.61 &0.66 &0.69 \\
        &SuperGlue &0.38 &0.72 &\textbf{0.86} &\textbf{0.92} &\textbf{0.94} &\textbf{0.96} &\textbf{0.96} &\textbf{0.97} &\textbf{0.97} &\textbf{0.97} \\
        &LightGlue &0.34 &0.68 &0.82 &0.89 &0.92 &0.94 &0.95 &0.95 &0.95 &0.96 \\
        &SAM  &0.37 &0.69 &0.82 &0.87 &0.90 &0.92 &0.93 &0.93 &0.93 &0.94 \\
        \rowcolor{table_bg}
		&\textbf{SceneGlue}  &\textbf{0.39} &\textbf{0.73} &\textbf{0.86} &0.90 &0.92 &0.93 &0.94 &0.94 &0.95 &0.95\\

		\midrule
        \multirow{11}{*}{Illumination change}
        &SIFT+CAPS &0.42 &0.59 &0.69 &0.78 &0.84 &0.88 &0.91 &0.93 &0.94 &0.95\\
        &D2Net &0.17 &0.36 &0.53 &0.67 &0.77 &0.84 &0.88 &0.90 &0.92 &0.93\\
        &SuperPoint &0.42 &0.58 &0.68 &0.74 &0.78 &0.80 &0.81 &0.82 &0.83 &0.83\\
        &SuperPoint+CAPS &0.42 &0.60 &0.73 &0.80 &0.85 &0.88 &0.90 &0.92 &0.93 &0.94\\
        &R2D2 &0.38 &0.66 &0.81 &0.87 &0.90 &0.91 &0.92 &0.92 &0.93 &0.93 \\
        &Patch2Pix &0.53 &\textbf{0.81} &\textbf{0.90} &\textbf{0.95} &\textbf{0.97} &\textbf{0.98} &0.98 &\textbf{0.99} &0.99 &0.99\\
        &NCNet &\textbf{0.78} &0.78 &0.78 &0.78 &0.79 &0.79 &0.79 &0.85 &0.90 &0.93\\
        &SuperGlue &0.46 &0.63 &0.75 &0.82 &0.86 &0.89 &0.90 &0.92 &0.93 &0.94 \\
        &LightGlue &0.49 &0.71 &0.86 &0.93 &0.96 &\textbf{0.98} &\textbf{0.99} &\textbf{0.99} &0.99 &0.99\\
        &SAM &0.50 &0.71 &0.84 &0.91 &0.95 &0.97 &0.98 &\textbf{0.99} &0.99 &0.99 \\
        \rowcolor{table_bg}
		&\textbf{SceneGlue}  &0.55 &0.77 &0.89 &0.94 &\textbf{0.97} &\textbf{0.98} &\textbf{0.99} &\textbf{0.99} &\textbf{1.00} &\textbf{1.00}\\

		\midrule
        \multirow{11}{*}{Overall}
		&SIFT+CAPS &0.39& 0.56& 0.65& 0.72& 0.76& 0.79& 0.81& 0.83& 0.84& 0.85 \\
		&D2Net &0.11 &0.27 &0.43 &0.56 &0.67 &0.73 &0.78 &0.81 &0.83 &0.84 \\
		&SuperPoint &0.33& 0.53& 0.64& 0.70 &0.73& 0.75& 0.77& 0.78& 0.79& 0.79 \\
		&SuperPoint+CAPS &0.34& 0.56& 0.69& 0.76& 0.80 & 0.83& 0.84& 0.86& 0.87& 0.88 \\
		&R2D2 &0.33& 0.62& 0.76& 0.81& 0.84& 0.85& 0.86& 0.87& 0.87& 0.87 \\
		&Patch2Pix &0.37& 0.64& 0.75& 0.81& 0.85& 0.87& 0.88& 0.89 & 0.90 & 0.91 \\
		&NCNet &0.38& 0.42& 0.46& 0.52& 0.57& 0.63& 0.67& 0.73& 0.78& 0.81 \\
		&SuperGlue &0.44& 0.71& 0.83& 0.89& 0.92& 0.93& 0.94& 0.95& 0.95& 0.96 \\
        &LightGlue &0.40 &0.68 &0.83 &0.90 &0.94 &0.95 &\textbf{0.96} &\textbf{0.97} &\textbf{0.97} &\textbf{0.97}\\
		&SAM &0.44& 0.72& 0.85& \textbf{0.91}& 0.94& 0.95& \textbf{0.96}&  0.96& 0.96& \textbf{0.97}\\
        \rowcolor{table_bg}
		&\textbf{SceneGlue} &\textbf{0.45}& \textbf{0.73}& \textbf{0.86}& \textbf{0.91}& \textbf{0.95}& \textbf{0.96}& \textbf{0.96}& \textbf{0.97}&  \textbf{0.97}& \textbf{0.97}\\ \bottomrule
	\end{tabular}	
	\label{tab:HPatches}
\end{table*}

\section{Experiments}
\label{experiments}
\subsection{Implementation Details}
Experimental studies on various applications based on feature matching are presented in this section, including image matching, homography estimation, outdoor pose estimation, and outdoor visual localization. Following convention, SceneGlue is trained on the Oxford100k dataset \cite{Oxford100k} for homography estimation experiments and on the MegaDepth dataset \cite{MegaDepth} for pose estimation and image matching experiments. The model is implemented by programming with Python under the framework of PyTorch. SceneGlue involves 9 parallel-attention layers, and all intermediate features have the same dimension $C=256$. The matching threshold $\theta$ is set to 0.2. For the homography estimation experiment, the AdamW optimizer \cite{adam} is employed for 10 epochs using the cosine decay learning rate scheduler and 1 epoch of linear warm-up. A batch size of 8 and an initial learning rate of 0.0001 are used. For the outdoor pose estimation experiment, the AdamW optimizer is utilized for 30 epochs using the same learning rate scheduler and linear warm-up. A batch size of 2 and an initial learning rate of 0.0001 are used. All experiments are conducted on a single NVIDIA GeForce RTX 2060 SUPER GPU, 16GB RAM, and Intel Core i7-10700K CPU.

\subsection{Image Matching}
\noindent\textbf{Dataset.}
HPatches \cite{hpatches} is a widely used image-matching benchmark that consists of 116 sequences with various challenges, including viewpoint changes and illumination changes. Following conventional works \cite{d2net}, eight unreliable scenes are excluded for fair comparison. Each scene is composed of 5 image pairs with groundtruth homography matrices.

\noindent\textbf{Baselines.}
Various baseline methods are taken into comparison study. The SceneGlue is compared with hardcrafted feature descriptor SIFT \cite{sift}, learned feature descriptors D2Net \cite{d2net}, SuperPoint \cite{superpoint} and R2D2 \cite{r2d2}, and recently proposed advanced matchers SuperGlue \cite{superglue}, SAM \cite{XiaoyongICCV2023}, LightGlue \cite{LightGlue}, Patch2Pix \cite{patch2pix}, NCNet \cite{NCNet}, and CAPS \cite{caps}.

\noindent\textbf{Metrics.}
Following existing works, the mean matching accuracy (MMA) is employed to evaluate the performance of the proposed approach on image matching. The MMA is defined as the average percentage of correct matches in image pairs under different matching error thresholds. To provide a comprehensive evaluation, the MMA is tested with multiple matching error thresholds in the interval between 1 pixel to 10 pixels.

\noindent\textbf{Results.}
Experimental result on image matching is reported in Table \ref{tab:HPatches}. In general, the MMA increases with the increase of the matching error thresholds and gradually approaches 1. In the experiments on overall dataset, SceneGlue achieves the highest MMA at all the thresholds, followed by LightGlue and SAM. In the experiments on large viewpoint change, SceneGlue achieves the highest MMA under small matching error thresholds (from 1px to 3px), and exhibits slightly inferior performance compared to SuperGlue under large matching error thresholds. In the experiments on illumination change, Patch2Pix and SceneGlue show superior performance compared to other methods. When the matching error threshold becomes large, SceneGlue outperforms other methods, including SuperGlue and Patch2Pix. Although existing works also utilize deep graph neural networks and Transformers to implicitly enhance the interaction among point-level feature descriptors, they do not fully investigate scene-level correspondence. The leading performance of the proposed SceneGlue demonstrates that incorporating the learned scene description is helpful in point-level feature matching.

\subsection{Homography Estimation}
\noindent\textbf{Dataset.}
For the homography estimation experiment, SceneGlue is trained on the Oxford100k dataset and then evaluated on the $\mathcal{R}$1M dataset \cite{r1m}. Since the $\mathcal{R}$1M dataset contains over a million images of Oxford and Paris, 500 images are randomly sampled from it for efficient evaluation. To perform self-supervised training, random groundtruth homographies are generated to get image pairs.
We resize the images to 640$\times$480 pixels and detect 512 keypoints in the image.
When the detected keypoints are not enough, random keypoints are added for efficient batching.

\noindent\textbf{Baselines.}
We employ the Nearest Neighbor (NN) matcher, NN matcher with outlier filtering methods \cite{pointcn,OANet}, and attention-based matcher SuperGlue \cite{superglue}, LightGlue \cite{LightGlue} and SAM \cite{XiaoyongICCV2023} as baseline matchers.
All matchers in Table \ref{tab:Homography} utilize SuperPoint \cite{superpoint} as input descriptors for a fair comparison.
The results of SuperGlue are from our implementation.

\noindent\textbf{Metrics.}
The groundtruth matches are computed from the generated homography and the keypoint coordinates of the two images.
We evaluate the precision and recall based on the groundtruth matches and further compute the F1-score.
We calculate reprojection error with the estimated homography and report the area under the cumulative error curve (AUC) up to 10 pixels.

\begin{figure*}[thb]
	\includegraphics[width=\linewidth]{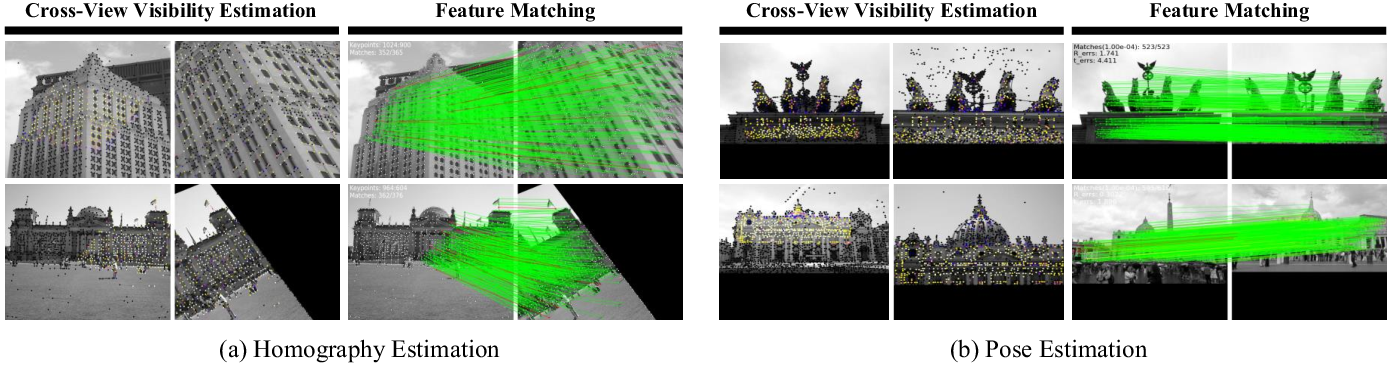}
	\centering
	 \caption{\textbf{Visualization of cross-view visibility estimation and feature matching results on (a) homography estimation and (b) outdoor pose estimation tasks.} SceneGlue precisely estimates the visible regions in cross-view images and further results in more robust and accurate point-level matching for both homography estimation and outdoor pose estimation.}
	 \label{fig:vis1}
\end{figure*}

\noindent\textbf{Results.}
Experimental results are reported in Table \ref{tab:Homography}, from which we can see that Transformer-based approaches show higher performance than the nearest neighbor matcher with outlier filtering algorithms. Although SuperGlue is also built on Transformer, its performances are $-2.36\%$ AUC below SceneGlue and -4.25$\%$ F1-score below SceneGlue. SceneGlue outperforms SAM by a noticeable gap in homography estimation. Compared to state-of-the-art outlier filtering methods OANet, SceneGlue achieves $+23.33\%$ improvement of F1-score. The results demonstrate that, despite the powerful modeling capability and global receptive field of the Transformer promoting the matching capacity, the way of organizing attention and scene-level guidance are also important factors that significantly strengthen matches in homography estimation. In addition, visualization of feature matching and learned scene descriptors are shown in Fig. \ref{fig:vis1}; visualization of the features aggregated by serial attention and parallel attention are presented in the supplementary files of this paper.

\begin{table}[t]
	\centering
    \caption{\textbf{Homography estimation on the $\mathcal{R}$1M dataset.} AUC @10 pixels is reported. The best method is highlighted in \textbf{bold}.}
	\label{tab:R1M_result}
	{	 \resizebox{\linewidth}{!}{
	\begin{tabular}{lcccc}
	\toprule
		Matcher                     & AUC (\%)                    & Precision (\%)                  & Recall (\%)                & F1-score (\%)          \\ \midrule
		NN                          & 39.47                   & 21.7                        & 65.4                   & 32.59              \\
		NN + mutual                 & 42.45                   & 43.8                        & 56.5                   & 49.35    \\
		NN + PointCN                & 43.02                   & 76.2                        & 64.2                   & 69.69    \\
		NN + OANet                  & 44.55                   & 82.8                        & 64.7                   & 72.64    \\
		SuperGlue                   & 51.94                   & 86.2    		            & 98.0 			         & 91.72    \\
        LightGlue                   & 51.75                   & 88.7                        & 98.1                   & 93.16    \\
		SAM                         & 53.80                   & 89.5                          &98.1                   & 93.64    \\
		\rowcolor{table_bg}
        \textbf{SceneGlue} 				& \textbf{54.30}   	  & \textbf{93.2}	   	    & \textbf{98.9} 		 & \textbf{95.97} \\ \bottomrule
	\end{tabular}	}}	\label{tab:Homography}
\end{table}

\subsection{Outdoor Pose Estimation}
\noindent\textbf{Dataset.}
To evaluate the performance of the proposed method on outdoor pose estimation, the model is trained on the MegaDepth dataset \cite{MegaDepth} that consists of 1M internet images of 196 different outdoor scenes. For training, 200 pairs of images in each scene are randomly sampled for each epoch. The images are resized to 960$\times$720 pixels and detect 1024 keypoints.
We follow the LoFTR to evaluate using two scenes, ``Sacre Coeur" and ``St. Peter's Square", which are excluded from the training set. The test set contains a total of 1500 image pairs. Besides the MegaDepth dataset, the trained model is also evaluated on the YFCC100M dataset \cite{2016YFCC100M}.
For evaluation, the YFCC100M image pairs and groundtruth poses provided by SuperGlue are used.

\noindent\textbf{Baselines.}
Various baseline methods are taken into comparison, including Transformer-based matchers SuperGlue \cite{superglue}, SAM \cite{XiaoyongICCV2023}, LightGlue \cite{LightGlue}, SGMNet \cite{SGMNet}, DenseGAP \cite{DenseGAP}, and ClusterGNN \cite{ClusterGNN}, as well as NN matcher with outlier filtering \cite{OANet}. Since the training source code of SuperGlue is unavailable, the results of SuperGlue are from our implementation. In addition, as one of the most widely recognized vision foundation models, DINOv2 \cite{oquab2023dinov2} is also taken into comparison. In our experiments, 1024 keypoints that use DINOv2 features are matched by the state-of-the-art matcher SuperGlue \cite{superglue} which is the baseline of this work.

\noindent\textbf{Metrics.}
The AUC of pose errors at the thresholds ($5^\circ$, $10^\circ$, $20^\circ$) are reported.
Both approximate AUC \cite{OANet} and exact AUC \cite{superglue} are evaluated on the YFCC100M dataset for a fair comparison.

\noindent\textbf{Results.}
Experimental results on the MegaDepth dataset and the YFCC100M dataset are reported in Table \ref{table:MegaDepth} and Table \ref{tab:YFCC100M}, respectively.

\begin{table}[t]
	\centering
	\small
	\caption{\textbf{Outdoor position estimation on the MegaDepth dataset.} AUCs of pose errors at the thresholds {@5$^\circ$}, {@10$^\circ$}, {@20$^\circ$} are reported. The best method is highlighted in \textbf{bold}.}
	{
\begin{tabular}{cclcccc}
			\toprule
            \multirow{2}{*}{Feature} &\multicolumn{1}{c}{\multirow{2}{*}{Matcher}} &\multicolumn{3}{c}{Pose estimation AUC (\%)} \\
            \cmidrule(l){3-5}
            & &{@$5^\circ$}  &{@$10^\circ$}  &{@$20^\circ$} \\
			\midrule
            DINOv2                          &SuperGlue &31.5 &40.8 &45.3\\
            \midrule
            \multirow{6}{*}{SuperPoint} 	&SuperGlue &42.2 &59.0 &73.6 \\
                                            &SGMNet &40.5 &59.0 &73.6 \\
											&DenseGAP &41.2 &56.9 &70.2 \\
											&ClusterGNN &\textbf{44.2} &58.5 &70.3 \\
											&LightGlue  &41.8  &\textbf{60.0}  &74.4 \\
											&SAM &40.9 &58.4 &72.3 \\
											\rowcolor{table_bg}
											&\textbf{SceneGlue}       &42.0 & \textbf{60.0} &\ \textbf{74.6}  \\

			\bottomrule
		\end{tabular}}\label{table:MegaDepth}
\end{table}

From Table \ref{table:MegaDepth}, we can see that although state-of-the-art approaches tend to utilize the global receptive field of the Transformer to leverage matching performance, SceneGlue shows competitive performance on outdoor pose estimation. Specifically, SceneGlue performs better than SGMNet, DenseGAP and SAM at three thresholds of $(5^{\circ}, 10^{\circ}, 20^{\circ})$. Although SuperGlue outperforms SceneGlue at the threshold of $5^{\circ}$, it does not continuously perform well at the thresholds of $10^{\circ}$ and $20^{\circ}$. As strong competitive method, LightGlue achieves the same performance with the proposed SceneGlue at only one threshold $10^\circ$, and is inferior at other two thresholds of $5^{\circ}$ and $20^{\circ}$. As another state-of-the-art approach, ClusterGNN achieves the highest AUC at threshold $5^{\circ}$, but it falls behind with the increase of the threshold. The proposed SceneGlue does not outperform the ClusterGNN because of the ClusterGNN uses a progressive clustering module to adaptively divide keypoints into different subgraphs to reduce redundant connectivity. The progressive clustering mechanism is beneficial to perceiving small pose changes. The ClusterGNN is therefore effective at small thresholds such as $5^{\circ}$. In particular, SceneGlue outperforms the state-of-the-art ClusterGNN by gaps of $+1.5\%$ AUC and $+4.3\%$ AUC at thresholds of $10^{\circ}$ and $20^{\circ}$, respectively. When using the features of the vision foundation model DINOv2 \cite{oquab2023dinov2}, SuperGlue shows poorer performance. DINOv2 is a family of foundation models producing universal features suitable for various visual tasks, \emph{e.g.}, image classification, instance retrieval, video understanding, depth estimation and semantic segmentation, that cover both image-level tasks and pixel-level tasks. Although it shows attractive performances on many tasks, it is not suitable for specific tasks like pose estimation that requires very precise recognition of feature correspondences.

Furthermore, one can find in Table \ref{tab:YFCC100M} that SceneGlue achieves the best performance at all thresholds in approximate AUC and one threshold in exact AUC on the YFCC100M dataset, demonstrating the superior capacity of SceneGlue in outdoor pose estimation. LightGlue demonstrates superior exact AUC performance at the $5^{\circ}$ and $10^{\circ}$ thresholds. This strength applies primarily to easy image pairs, whereas it underperforms compared to SceneGlue on challenging pairs with significant scale variations. Compared to the attention-based matcher SuperGlue, SceneGlue achieves $(+3.47\%, +4.23\%, +3.86\%)$ improvement on exact AUC and $(+5.55\%, +4.81\%, +4.29\%)$ improvement on approximate AUC at three thresholds of $(5^{\circ}, 10^{\circ}, 20^{\circ})$, respectively. The matching results by SceneGlue, SAM, and the SuperGlue on homography estimation and outdoor pose estimation tasks are visualized in the supplementary files of this paper.

\begin{table}[t]
	\centering
  	\caption{\textbf{Pose estimation on YFCC100M dataset.} The proposed method outperforms other methods at all thresholds.}{
	\begin{tabular}{lcccccc}
	\toprule
	\multirow{2}{*}{Matcher} & \multicolumn{3}{c}{Exact AUC (\%)} & \multicolumn{3}{c}{Approx. AUC (\%)} \\ \cmidrule(l){2-7}
							 & @5$^{\circ}$       & @10$^{\circ}$      & @20$^{\circ}$     & @5$^{\circ}$        & @10$^{\circ}$      & @20$^{\circ}$       \\ \midrule
	NN + mutual              & 16.94    & 30.39    & 45.72   & 35.00     & 43.12    & 54.25    \\
	NN + OANet               & 26.82    & 45.04    & 62.17   & 50.94     & 61.41    & 71.77    \\
	SuperGlue                & 28.45    & 48.6     & 67.19   & 55.67     & 66.83    & 74.58    \\
    LightGlue                & \textbf{32.34}    & \textbf{52.98}    & 70.85   & 60.58     & 71.20    & 78.37\\
    SAM                      &31.45     &52.27     &70.31    &60.65      &70.95     &78.03    \\
	\rowcolor{table_bg}
	\textbf{SceneGlue}            & {31.92}    & {52.83}    & \textbf{71.05}   & \textbf{61.22}     & \textbf{71.64}    & \textbf{78.87}    \\ \bottomrule
	\end{tabular}%
	}\label{tab:YFCC100M}
\end{table}

\subsection{Indoor Pose Estimation}
\noindent\textbf{Dataset.}
The indoor pose estimation experiments are conducted on ScanNet \cite{scannet} and InLoc \cite{inloc}. The ScanNet \cite{scannet} is a large-scale indoor dataset that provides monocular image sequences along with ground-truth camera poses and depth maps. It offers clearly defined training, validation, and testing splits, each covering distinct scenes. Following SuperGlue \cite{superglue} and SGMNet \cite{SGMNet}, 1500 test pairs are utilized for evaluation.

The InLoc dataset \cite{inloc} consists of RGB query images captured with a smartphone camera in indoor environments. The corresponding 6-DoF reference poses were manually verified for evaluations.
Following DiffGlue \cite{DiffGlue}, the images of two floors, DUC1 and DUC2, are used for testing. For each query image, 40 reference images are retrieved, and feature matching is performed using up to 4096 keypoints detected by SuperPoint \cite{superpoint}. The camera pose is then estimated using RANSAC and a PnP solver.

\noindent\textbf{Baselines.}
Various baseline methods are taken into comparison, including handcrafted method AdaLAM \cite{AdaLAM}, learned matchers SuperGlue \cite{superglue}, SAM \cite{XiaoyongICCV2023}, LightGlue \cite{LightGlue}, SGMNet \cite{SGMNet}, DiffGlue \cite{DiffGlue} and ResMatch \cite{resmatch}, as well as NN matcher with outlier filtering \cite{OANet}. Since the training source code of SuperGlue is unavailable, the results of SuperGlue are from our implementation.

\noindent\textbf{Metrics.}
On the ScanNet dataset \cite{scannet}, the AUC of pose errors at the thresholds ($5^\circ$, $10^\circ$, $20^\circ$) are reported. On the InLoc dataset \cite{inloc}, we follow DiffGlue to report pose recall at multiple thresholds of distance and orientation: (0.25, $10^\circ$) / (0.5m, $10^\circ$) / (1.0m, $10^\circ$).

\noindent\textbf{Results.}
From the results reported in Table \ref{table:ScanNet}, we can find that SceneGlue attains the best results under rigorous evaluation conditions (pose errors at the thresholds of $5^\circ$ and $10^\circ$). When the experimental conditions are relatively relaxed, LightGlue slightly outperforms SceneGlue. From the results reported in Table \ref{table:InLoc}, one can see that SuperGlue, DiffGlue and SceneGlue show better performance than other methods, as SceneGlue and SuperGlue respectively attain 3 best results, and DiffGlue attains 2 best results. The experimental results show that SceneGlue achieves more accurate indoor pose estimation under challenging conditions, indicating its higher precision.

\begin{table}[t]
	\centering
	\small
	\caption{\textbf{Indoor position estimation on the ScanNet dataset.} The best method is highlighted in \textbf{bold}.}
	{
\begin{tabular}{c c l c c c c}
			\toprule
            \multicolumn{1}{c}{\multirow{2}{*}{Matcher}} &\multicolumn{3}{c}{Pose estimation AUC (\%)} \\ \cmidrule(l){2-4}
                     &{@$5^\circ$}  &{@$10^\circ$}  &{@$20^\circ$} \\
			\midrule
            AdaLAM	&6.72	&15.82	&27.37  \\
            OANet	&10.04	&25.09	&38.01  \\
            SGMNet	&13.06	&28.94	&46.32  \\
            SuperGlue	&13.95	&29.48	&46.07  \\
            LightGlue	&14.00	&29.75	&\textbf{46.77}  \\
            SAM	    &13.95	&28.50	&44.32  \\
            \rowcolor{table_bg}
            SceneGlue	&\textbf{14.54}	&\textbf{30.00}	&46.64  \\
			\bottomrule
		\end{tabular}}\label{table:ScanNet}
\end{table}

\subsection{Outdoor Visual Localization}
\noindent\textbf{Dataset.}
Outdoor visual localization is an important application of feature matching. This application is experimentally studied to further test the performance of the proposed method. In particular, we evaluate long-term visual localization on the large-scale  Aachen Day-Night (v1.0 and v1.1) benchmarks \cite{Aachenset} that provide various challenging conditions with images from the old inner city of Aachen. The key challenge lies in matching images with extreme day-night changes for 824 daytime and 98 nighttime queries. The HLoc toolbox \cite{HLoc} is utilized for mapping
and localization and report the accuracy at multiple error thresholds.

\noindent\textbf{Baselines.}
The baselines comprise simple matchers such as mutual Nearest Neighbor (MNN) matcher. Also, filtering-based methods, including OANet \cite{OANet} and AdaLAM \cite{AdaLAM} are also included. The final part is Transformer-based matchers such as SuperGlue \cite{superglue}, SAM \cite{XiaoyongICCV2023}, LightGlue \cite{LightGlue} and SGMNet \cite{SGMNet}.

\noindent\textbf{Metrics.}
The evaluation is conducted by using the Visual Localization Benchmark, which takes a pre-defined visual localization pipeline based on COMLAP. The successful localized images are counted within three error tolerances (0.25m, 2$^\circ$) / (0.5m, 5$^\circ$) / (5m, 10$^\circ$), representing the maximum position error in meters and degrees.

\noindent\textbf{Results.}
Experimental results on outdoor visual localization are reported in Table \ref{tab:aachen}, from which one can find the best results are not as concentrated as other experiments. Nevertheless, SceneGlue achieves the highest performance in day sequences at error thresholds $(0.25m,2^\circ)$ and $(0.5m,5^\circ)$ on the v1.0 dataset, in night sequences at the error threshold $(5m,10^\circ)$ on the v1.0 dataset and at all the error thresholds on the v1.1 dataset. Overall, SceneGlue achieves 6 best results (bold in Table \ref{tab:aachen}) on the two datasets, followed by SuperGlue that achieves 5 best results and LightGlue that achieves 2 best results.

\begin{table}[t]
	\centering
	\small
	\caption{\textbf{Indoor position estimation on the InLoc dataset.} The best method is highlighted in \textbf{bold}.}
	{
    \begin{tabular}{ccc}
			\toprule
            \multicolumn{1}{c}{\multirow{2}{*}{Matcher}} &\multicolumn{2}{c}{(0.25, $10^\circ$) / (0.5m, $10^\circ$) / (1.0m, $10^\circ$)} \\ \cmidrule(l){2-3}
                     &DUC1   &DUC2 \\
			\midrule	
            SuperGlue	&44.9 / 66.2 / \textbf{78.8}	&46.6 / \textbf{74.0} / \textbf{77.1}\\
            SGMNet	&39.9 / 56.6 / 70.2	&39.7 / 59.5 / 65.6\\
            ResMatch	&42.9 / 61.6 / 73.7	&38.2 / 62.6 / 69.5\\
            LightGlue	&44.4 / 64.1 / 75.8	&42.7 / 67.9 / 73.3\\
            DiffGlue	&46.5 / \textbf{67.2} / 78.3	&\textbf{49.6} / 71.8 / 76.3\\
            \rowcolor{table_bg}
            SceneGlue	&\textbf{47.5} / 66.7 / \textbf{78.8}	&45.0 / 63.4 / \textbf{77.1}\\
			\bottomrule
		\end{tabular}}\label{table:InLoc}
\end{table}

\begin{table}
	\centering
	\caption{\textbf{Results (\%) of outdoor visual localization on Aachen dataset}. The \textbf{best} results are highlighted in bold.}
	\begin{tabular}{ll|cc}
			\toprule
			& \multirow{2}{*}{Matcher} & Day & Night \\
			&  & \multicolumn{2}{c}{$(0.25m,2^\circ) / (0.5m,5^\circ) / (5m,10^\circ)$}\\
			\midrule
\multirow{9}{*}{v1.0}
			& MNN & 85.4 / 93.3 / 97.2 & 75.5 / 86.7 / 92.9 \\
			& OANet & - & 77.6 / 86.7 / 98.0\\
			& AdaLAM & - & 78.6 / 86.7 / 98.0 \\			
			& SuperGlue & 88.0 / 94.9 / \textbf{98.4} & 82.7 / \textbf{92.9} / \textbf{100.0}			\\
            & LightGlue & 88.0 / 93.8 / 97.5          & \textbf{84.7} / 91.8 / 99.0       \\
			& SGMNet & 86.8 / 94.2 / 97.7 & {83.7} / 91.8 / 99.0 \\
			& SAM & 87.6 / 93.3 / 97.5			&{83.7} / 91.8 / 99.0			\\
            \rowcolor{table_bg}
			& \textbf{SceneGlue} & \textbf{88.8} / \textbf{95.0} / 98.2			&{83.7} / 92.8 / \textbf{100.0}			\\
			
			\midrule
\multirow{7}{*}{v1.1}
			& MNN & 87.9 / 93.6 / 96.8 & 70.2 / 84.8 / 93.7 \\
			& AdaLAM & - & 73.3 / 86.9 / 97.9 \\
			& SuperGlue & \textbf{89.8} / 95.9 / \textbf{99.2} & 75.4 / 89.5 / 99.0	\\
            & LightGlue &89.6 / 95.8 / \textbf{99.2}                    & 72.8 / 88.0 / 99.0 \\
			& SGMNet & 88.7 / \textbf{96.2} / 98.9             & 75.9 / 89.0 / 99.0 \\
			& SAM & 82.8 / 93.7 / 98.4			& 73.3 / 90.1 / 97.9			\\
            \rowcolor{table_bg}
			& \textbf{SceneGlue} & 89.2 / 95.9 / 99.0			& \textbf{77.5} / \textbf{91.1} / \textbf{99.5}			\\
			\bottomrule			
		\end{tabular}
	\label{tab:aachen}
\end{table}

\subsection{Ablation Study}
To make deeper analyses of each contribution made by the proposed SceneGlue, comprehensive ablation experiments are conducted to test the validity of every proposal. In specific, the proposed informative feature representation, parallel attention, and cross-view visibility estimation modules presented in Section \ref{sec:Methodology} are experimentally studied separately. As reported in Table \ref{tab:ablation}, the ablation studies are composed of five sets of experiments:
\begin{itemize}
  \item[i)] The baseline framework SuperGlue that utilizes serial attention is first tested as a basic reference.
  \item[ii)] The MLP-PE is replaced to the Wave-PE that is presented in Section \ref{Wave position encoder}.
  \item[iii)] The serial attention is replaced to the parallel attention proposed in Section \ref{sec:Parallelattention}.
  \item[iv)] The proposed informative feature representation is applied to SceneGlue without cross-view visibility estimation, to show the effectiveness of the feature representation method itself.
  \item[v)] To test the contribution of the proposed cross-view visibility estimation in Section \ref{sec:Visibility}, it is finally applied to SceneGlue.
\end{itemize}

\begin{table}[tb]
	\centering
  	\caption{\textbf{Ablation study on each proposal of SceneGlue using the $\mathcal{R}$1M dataset.} The \textbf{best} results are highlighted in bold.}
  	\label{tab:ablation}
	\resizebox{\linewidth}{!}{%
	\begin{tabular}{lccc}
	\toprule
	Proposal  & Precision (\%) & Recall (\%) & F1-score (\%) \\ \midrule
	(i) Serial attention      & 86.2     & 98.0   & 91.72    \\
	(ii) Wave-PE              & 86.8     & 98.9   & 92.46    \\
	(iii) Parallel attention      & 88.8     & 98.8  & 93.50 (\textcolor{green}{+1.78})    \\
	(iv) + Multi-scale feature network      & 91.0     & \textbf{99.0}  & 94.89 (\textcolor{green}{+1.39})    \\
	\rowcolor{table_bg}
	(v) \textbf{+ Visibility estimation (full)}       & \textbf{93.2} (\textcolor{green}{\textbf{+2.20}})     &98.9  & \textbf{95.97} (\textcolor{green}{\textbf{+1.08}})    \\ \bottomrule
	\end{tabular}}
\end{table}

\noindent\textbf{Wave-PE.}
To test effectiveness of the Wave-PE, it is added to the baseline SuperGlue. As can be seen in Table \ref{tab:ablation}, both precision and recall are improved by the Wave-PE. The performance is leveraged by $+0.74\%$ F1-score after replacing the MLP-PE with the proposed Wave-PE. One can therefore conclude that the Wave-PE incorporates more useful information to the descriptors than MLP-PE.

\noindent\textbf{Parallel Attention.}
Despite the great successful attempts made by SuperGlue-like methods, they calculate self- and cross-attention in a serial manner that does not fully explore the potential of deep parallel computing. By comparing the results of experiments i) and iii) in Table \ref{tab:ablation}, one can find that both precision and recall are improved by the parallel attention because self- and cross-attentions are arranged in a parallel manner which achieves a win-win performance in terms of accuracy and efficiency.

\noindent\textbf{Multi-Scale Feature Network.}
In theory, the proposed informative feature representation enhances existing matchers in two aspects: a) the designed lightweight multi-scale feature representation makes descriptor more robust to scale-varying scenarios; b) benefiting from the proposed wave position encoding, the relationship between descriptor and position can be dynamically adjusted by amplitude and phase, which leads to better fusion of position information. By comparing the results of experiments i), ii) and iv) in Table \ref{tab:ablation}, one can find that the F1-score is significantly increased by $+1.39\%$ improvement after adding the multi-scale feature network. The results prove theoretical analyses.

\noindent\textbf{Cross-View Visibility Estimation.}
As a key contribution in obtaining scene-level information for matching point-level features, the commonly visible area in images taken from different views is estimated in Section \ref{sec:Visibility}. We can see from Table \ref{tab:ablation} that the matching performance measured by F1-score is inspiringly leveraged by $+1.08\%$ improvement when introducing the cross-view visibility estimation module. Benefiting from the delicate design that assigns the multi-scale local descriptors to the commonly-visible area or commonly-invisible area through a $\mathrm{Softmax}$ classifier based on the Transformer architecture, the learned scene descriptors provide rich information about the commonly-visible area between different views.

\noindent\textbf{Hyper-Parameter $\alpha$.}
To test the effect of the importance of each loss term, the hyper-parameter $\alpha$ defined in the loss function is experimentally studied. As reported in Table \ref{table:alpha}, one can find that the performance is improved with the increase of $\alpha$, indicating that the scene-awareness term is important for the matching task. as the best performance is achieved when $\alpha = 8$, we set $\alpha$ to 8 in other experiments.
\begin{table}[t]
	\centering
  \caption{\textbf{Ablation study on the hyper-parameter $\alpha$ using the $\mathcal{R}$1M dataset.} The \textbf{best} result is highlighted in bold.}
	\resizebox{0.95\linewidth}{!}{%
	\begin{tabular}{cccccccc}
	\toprule
	$\alpha$  & 0 & 0.5 & 1 & 2 & 4 & 8 & 10 \\ \midrule
    Precision (\%)    &91.90  &92.30  &92.20 &92.60 &92.90 &\textbf{93.20} &93.20 \\
    Recall (\%)       &98.40  &98.70  &98.80 &98.90 &98.90 &\textbf{98.90} &98.90 \\
    F1-Score (\%)     &95.03  &95.40  &95.42 &95.62 &95.78 &\textbf{95.97} &95.96  \\ \bottomrule
	\end{tabular}	}
	\label{table:alpha}
\end{table}

\noindent\textbf{Number of Parameters of the Wave-PE.}
To analysis the contribution of the Wave-PE, it is compared with the MLP-PE in Table \ref{table:WaveNum}. One can see that the Wave-PE holds 1.35M parameters in our implementation. A simple MLP-PE can use very little parameters, but its performance is low. Even using a larger number of parameters, the MLP-PE does not achieve better performance. When Wave-PE and MLP-PE hold similar number of parameters (1.35M and 1.42M), the former method shows higher precision, recall and F1-score.

\begin{table}[t]
	\centering
  \caption{\textbf{Ablation study on the number of parameters of the Wave-PE using the $\mathcal{R}$1M dataset.} The \textbf{best} result is highlighted in bold.}
	\begin{tabular}{cccc}
	\toprule
	Method              &Precision (\%) &Recall (\%) & F1-score (\%) \\ \midrule
	(i) MLP-PE (0.11M)    &86.2       &98.0  &91.72 \\
    (ii) MLP-PE (1.42M)   &86.6       &98.5  &92.17 \\
    (iii) Wave-PE (1.35M) &\textbf{86.8}       &\textbf{98.9}  &\textbf{92.46} \\ \bottomrule
	\end{tabular}
	\label{table:WaveNum}
\end{table}

\begin{figure*}[thb]
	\includegraphics[width=\linewidth]{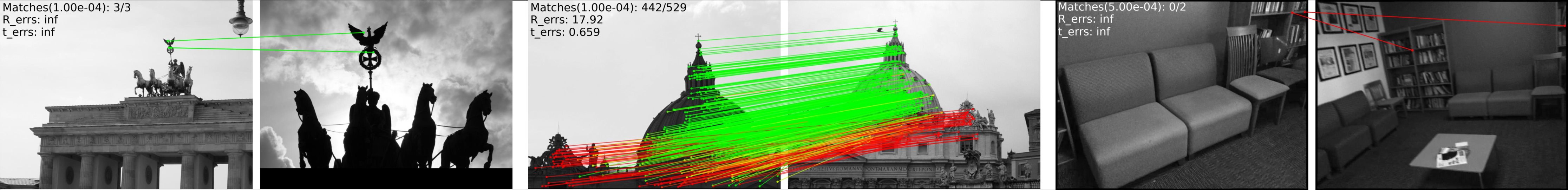}
	\centering
	 \caption{Failure cases.}
	 \label{fig:failure}
\end{figure*}

\noindent\textbf{Number of Parameters of the Multi-Scale Feature Network.}
To analysis the effect of the number of parameters of the multi-scale feature network, an ablation study is performed by changing the number of parameters through changing layers but keeping neural network structure same. Experimental results reported in in Table \ref{table:featurepara} show that the best performance is achieved when the number of parameters is around 222K. Therefore, we can draw the conclusion that too much or too little parameters in the multi-scale feature network do not yield better performance.

\begin{table}[t]
	\centering
  \caption{\textbf{Ablation study on the number of parameters of the multi-scale feature network using the $\mathcal{R}$1M dataset.} The \textbf{best} result is highlighted in bold.}
	\begin{tabular}{cccc}
	\toprule
	Parameters      &Precision (\%) &Recall (\%)    & F1-score (\%) \\ \midrule
    55K             &93.0       &98.7       &95.76 \\
    111K            &92.9       &98.8       &95.78 \\
    222K            &\textbf{93.2}       &\textbf{98.9}       &\textbf{95.97} \\
    444K            &93.0       &98.8       &95.84 \\
    888K            &92.9       &\textbf{98.9}       &95.83 \\ \bottomrule
	\end{tabular}
	\label{table:featurepara}
\end{table}

\subsection{Efficiency Analysis}
To analyze the efficiency of the proposed SceneGlue, its parameter number, floating point operations (FLOPs) and runtime are reported in Table \ref{ta:Efficiency}. Other state-of-the-art feature matching approaches, including both sparse matchers SuperGlue \cite{superglue}, SAM \cite{XiaoyongICCV2023}, LightGlue \cite{LightGlue}, SGMNet \cite{SGMNet} and dense matchers LoFTR \cite{loftr} and ASpanFormer \cite{ASpanFormer} are also cited for references. To make a fair comparison, all the sparse methods are evaluated using a typical input of 2048 keypoints. The dense methods are evaluated using a typical input resolution of $800 \times 800$ pixels.

One can find in Table \ref{ta:Efficiency} that SceneGlue involves the least parameters compared with other methods. Through comparing SceneGlue and SAM, one can find that SceneGlue demonstrates significantly better performance than SAM while utilizing fewer parameters. Although the FLOPs of SceneGlue are slightly higher than those of SAM, the increments remain well within a reasonable and acceptable range. Other sparse matchers, including SuperGlue, LightGlue and SGMNet, utilize more learnable parameters and involves more FLOPs than SceneGlue. Sparse methods are much faster than dense ones in general, as LoFTR and ASpanFormer take $3\sim5$ times of runtime of SceneGlue. SuperGlue is a little faster than SceneGlue, but the gap is almost insignificant. In summary, considering both the efficiency and the superior performances of SceneGlue on various tasks, there's a reasonable conclusion that SceneGlue achieves leading performance in feature matching.

\section{Limitation and Discussion}
\label{Limitation}
Since the objective of the proposed SceneGlue is to establish pixel-level correspondences between cross-view images, the model is trained without semantic annotation supervision. In our implementation, only keypoint matching groundtruth is used as supervision during the training stage. Although SceneGlue demonstrates strong performance in keypoint matching, it lacks explicit semantic awareness, as its scene awareness is obtained by predicting overlapping regions between two images rather than learning semantic concepts. As illustrated by the failure cases in Fig. \ref{fig:failure}, mismatches may still occur when the illumination conditions or viewpoints between two images differ significantly. While scene awareness is not equivalent to semantic understanding, it substantially improves feature matching performance. We believe that, if appropriate semantic supervision is incorporated in more complex applications, SceneGlue could learn to align semantic regions more effectively, which may further enhance its robustness under extreme illumination or viewpoint variations.

\begin{table}[t]
	\centering
	\caption{\textbf{Efficiency analysis}.}
	\begin{tabular}{lccc}
	\toprule
	Methods  & \#Params (M) &FLOPs (G) & Runtime (ms)  \\ \midrule
    LoFTR     & 11.6   &753.66  & 373.50      \\
	ASpanFormer   & 15.8  &815.91   & 500.28      \\
	SuperGlue    & 12.0   &22.51  & 104.37               \\
    LightGlue &11.9 &34.89 &87.4      \\
    SGMNet &31.1 &30.10 &92.1      \\
	SAM   &14.8 &17.61  &111.42      \\ \bottomrule
	\rowcolor{table_bg}
	SceneGlue    & 11.2  &18.12   & 116.31      \\ \bottomrule
	\end{tabular}
	\label{ta:Efficiency}
\end{table}

\section{Conclusion}
\label{conclusion}
In this paper, we propose SceneGlue, an attention-based local feature matching framework that introduces scene-level awareness into point-level correspondence estimation. The method improves feature representation by sampling keypoints from multiple resolutions and fusing them through lightweight linear layers. In addition, SceneGlue performs self-attention and cross-attention in parallel, enabling simultaneous scene reasoning and more efficient computation. To further incorporate scene-level information, we design a Visibility Transformer that explicitly predicts cross-view visible regions, establishing scene-level correspondence between images. Notably, this scene awareness is learned without requiring additional scene-level annotations. Experimental results demonstrate that the proposed design significantly improves local feature matching performance across multiple tasks, highlighting the importance of integrating scene-level understanding into feature correspondence learning.
\section*{Acknowledgments}
The authors would like to thank editors and anonymous reviewers for their insightful comments and suggestions.

\ifCLASSOPTIONcaptionsoff
  \newpage
\fi



\bibliographystyle{IEEEtran}

%

\bibliography{egbib}

@article{2016YFCC100M,
  title   = {{YFCC100M}: The New Data in Multimedia Research},
  author  = { Thomee, B.  and  Elizalde, B.  and  Shamma, D. A  and  Ni, K.  and  Friedland, G.  and  Poland, D.  and  Borth, D.  and  Li, L. J. },
  journal={Commun. ACM},
  pages   = {64-73},
  year    = {2016}
}

@inproceedings{caps,
  author    = {Wang, Q.
               and Zhou, X.
               and Hariharan, B.
               and Snavely, N.},
  title     = {Learning Feature Descriptors Using Camera Pose Supervision},
booktitle = {Proc. Eur. Conf. Comput. Vis. (ECCV)},
  year      = {2020},
  pages     = {757--774}
}

@inproceedings{d2net,
  author    = {Dusmanu, M. and Rocco, I. and Pajdla, T. and Pollefeys, M. and Sivic, J. and Torii, A. and Sattler, T.},
  title     = {{D2-Net}: A Trainable {CNN} for Joint Description and Detection of Local Features},
  booktitle={Proc. IEEE/CVF Conf. Comput. Vis. Pattern Recognit. (CVPR)},
  year      = {2019},
  pages     = {8092-8101}
}

@inproceedings{hpatches,
  author    = {Balntas, V. and Lenc, K. and Vedaldi, A. and Mikolajczyk, K.},
  title     = {{HPatches}: A Benchmark and Evaluation of Handcrafted and Learned Local Descriptors},
  booktitle={Proc. IEEE/CVF Conf. Comput. Vis. Pattern Recognit. (CVPR)},
  pages     = {5173--5182},
  year      = {2017}
}

@inproceedings{loftr,
  author    = {Sun, J. and Shen, Z. and Wang, Y. and Bao, H. and Zhou, X.},
  title     = {{LoFTR}: Detector-Free Local Feature Matching With Transformers},
  booktitle={Proc. IEEE/CVF Conf. Comput. Vis. Pattern Recognit. (CVPR)},
  year      = {2021},
  pages     = {8922-8931}
}

@inproceedings{matchformer,
  title     = {{MatchFormer}: Interleaving attention in {Transformers} for feature matching},
  author    = {Wang, Q. and Zhang, J. and Yang, K. and Peng, K. and Stiefelhagen, R.},
  booktitle={Proc. Asia. Conf. Comput. Vis. (ACCV)},
  pages     = {2746--2762},
  year      = {2022}
}

@inproceedings{MegaDepth,
  author    = {Li, Z. and Snavely, N.},
  booktitle={Proc. IEEE/CVF Conf. Comput. Vis. Pattern Recognit. (CVPR)},
  title     = {{MegaDepth}: Learning Single-View Depth Prediction from Internet Photos},
  year      = {2018},
  pages     = {2041-2050}}

@inproceedings{oetr,
  title     = {Guide Local Feature Matching by Overlap Estimation},
  author    = {Chen, Y. and Huang, D. and Xu, S. and Liu, J. and Liu, Y.},
  booktitle = {Proc. AAAI Conf. Artif. Intell. (AAAI)},
  pages     = {365--373},
  year      = {2022}
}

@inproceedings{orb,
  author    = {Rublee, E. and Rabaud, V. and Konolige, K. and Bradski, G.},
  booktitle={Proc. Int. Conf. Comput. Vis. (ICCV)},
  title     = {{ORB}: An efficient alternative to {SIFT or SURF}},
  year      = {2011},
  pages     = {2564-2571},
}

@inproceedings{patch2pix,
  author    = {Zhou, Q. and Sattler, T. and Leal-Taixe, L.},
  title     = {{Patch2Pix}: Epipolar-Guided Pixel-Level Correspondences},
  booktitle={Proc. IEEE/CVF Conf. Comput. Vis. Pattern Recognit. (CVPR)},
  year      = {2021},
  pages     = {4669-4678}}

@inproceedings{pointcn,
  author    = {Yi, K. M. and Trulls, E. and Ono, Y. and Lepetit, V. and Salzmann, M. and Fua, P.},
  booktitle={Proc. IEEE/CVF Conf. Comput. Vis. Pattern Recognit. (CVPR)},
  title     = {{Learning to Find Good Correspondences}},
  year      = {2018},
  pages     = {2666-2674}}

@inproceedings{r1m,
  author    = {Radenovi\'{c}, F. and Iscen, A. and Tolias, G. and Avrithis, Y. and Chum, O.},
  title     = {{Revisiting Oxford and Paris}: Large-Scale Image Retrieval Benchmarking},
  booktitle={Proc. IEEE/CVF Conf. Comput. Vis. Pattern Recognit. (CVPR)},
  pages     = {5706--5715},
  year      = {2018}}

@inproceedings{r2d2,
  author    = {Revaud, J. and De Souza, C. and Humenberger, M. and Weinzaepfel, P.},
  booktitle={Proc. Adv. Neural Inf. Process. Syst. (NeurIPS)},
  title     = {{R2D2}: Reliable and Repeatable Detector and Descriptor},
  pages={12414--12424},
  year={2019}
}

@inproceedings{superglue,
  author    = {Sarlin, P. and DeTone, D. and Malisiewicz, T. and Rabinovich, A.},
  booktitle={Proc. IEEE/CVF Conf. Comput. Vis. Pattern Recognit. (CVPR)},
  title     = {{SuperGlue}: Learning Feature Matching With Graph Neural Networks},
  year      = {2020},
  pages     = {4937-4946}}

@inproceedings{superpoint,
  author    = {DeTone, D. and Malisiewicz, T. and Rabinovich, A.},
  booktitle={Proc. IEEE/CVF Conf. Comput. Vis. Pattern Recognit. Workshops (CVPRW)},
  title     = {{SuperPoint}: Self-Supervised Interest Point Detection and Description},
  year      = {2018},
  pages     = {224-236}}

@inproceedings{swin,
  title     = {Swin {Transformer}: Hierarchical vision {Transformer} using shifted windows},
  author    = {Liu, Z. and Lin, Y. and Cao, Y. and Hu, H. and Wei, Y. and Zhang, Z. and Lin, S. and Guo, B.},
  booktitle={Proc. Int. Conf. Comput. Vis. (ICCV)},
  pages     = {10012--10022},
  year      = {2021}
}

@INPROCEEDINGS{Aachenset,
  author={Sattler, Torsten and Maddern, Will and Toft, Carl and Torii, Akihiko and Hammarstrand, Lars and Stenborg, Erik and Safari, Daniel and Okutomi, Masatoshi and Pollefeys, Marc and Sivic, Josef and Kahl, Fredrik and Pajdla, Tomas},
  booktitle={Proc. IEEE/CVF Conf. Comput. Vis. Pattern Recognit. (CVPR)},
  title={Benchmarking {6DOF} Outdoor Visual Localization in Changing Conditions},
  year={2018},
  pages={8601-8610}
}

@INPROCEEDINGS{HLoc,
  author={Sarlin, Paul-Edouard and Cadena, Cesar and Siegwart, Roland and Dymczyk, Marcin},
  booktitle={Proc. IEEE/CVF Conf. Comput. Vis. Pattern Recognit. (CVPR)},
  title={From Coarse to Fine: Robust Hierarchical Localization at Large Scale},
  year={2019},
  pages={12708-12717}}

@inproceedings{inloc,
  title={{InLoc}: Indoor visual localization with dense matching and view synthesis},
  author={Taira, Hajime and Okutomi, Masatoshi and Sattler, Torsten and Cimpoi, Mircea and Pollefeys, Marc and Sivic, Josef and Pajdla, Tomas and Torii, Akihiko},
  booktitle={Proc. IEEE/CVF Conf. Comput. Vis. Pattern Recognit. (CVPR)},
  pages={7199--7209},
  year={2018}
}

@INPROCEEDINGS{SGMNet,
    author    = {Chen, Hongkai and Luo, Zixin and Zhang, Jiahui and Zhou, Lei and Bai, Xuyang and Hu, Zeyu and Tai, Chiew-Lan and Quan, Long},
    title     = {Learning To Match Features With Seeded Graph Matching Network},
  booktitle={Proc. Int. Conf. Comput. Vis. (ICCV)},
    year      = {2021},
    pages     = {6301--6310}}

@INPROCEEDINGS{DenseGAP,
  author={Kuang, Zhengfei and Li, Jiaman and He, Mingming and Wang, Tong and Zhao, Yajie},
  booktitle={Proc. Int. Conf. Pattern Recognit. (ICPR)},
  title={{DenseGAP}: Graph-Structured Dense Correspondence Learning with Anchor Points},
  year={2022},
  pages={542-549}}

@INPROCEEDINGS{ClusterGNN,
  author={Shi, Yan and Cai, Jun-Xiong and Shavit, Yoli and Mu, Tai-Jiang and Feng, Wensen and Zhang, Kai},
  booktitle={Proc. IEEE/CVF Conf. Comput. Vis. Pattern Recognit. (CVPR)},
  title={{ClusterGNN}: Cluster-based Coarse-to-Fine Graph Neural Network for Efficient Feature Matching},
  year={2022},
  pages={12507-12516}}

@article{sift,
title = {Distinctive Image Features from Scale-Invariant Keypoints},
journal = {Int. J. Comput. Vis.},
volume = {60},
number = {2},
pages = {91--110},
year = {2004},
author = {Lowe, David G.}}

@ARTICLE{OANet,
  author={Zhang, Jiahui and Sun, Dawei and Luo, Zixin and Yao, Anbang and Chen, Hongkai and Zhou, Lei and Shen, Tianwei and Chen, Yurong and Quan, Long and Liao, Hongen},
  journal={IEEE Trans. Pattern Anal. Mach. Intell.},
  title={{OANet}: Learning Two-View Correspondences and Geometry Using Order-Aware Network},
  year={2022},
  volume={44},
  number={6},
  pages={3110-3122}}

@ARTICLE{NCNet,
  author={Rocco, Ignacio and Cimpoi, Mircea and Arandjelovi\'{c}, Relja and Torii, Akihiko and Pajdla, Tomas and Sivic, Josef},
  journal={IEEE Trans. Pattern Anal. Mach. Intell.},
  title={{NCNet}: Neighbourhood Consensus Networks for Estimating Image Correspondences},
  year={2022},
  volume={44},
  number={2},
  pages={1020-1034}}

@INPROCEEDINGS{XiaoyongICCV2023,
  author={Xiaoyong Lu and Yaping Yan and Tong Wei and Songlin Du},
  booktitle={Proc. Int. Conf. Comput. Vis. (ICCV)},
  title={Scene-Aware Feature Matching},
  year={2023},
  pages={3704-3710}}

@INPROCEEDINGS{ASLFeat,
  author={Luo, Zixin and Zhou, Lei and Bai, Xuyang and Chen, Hongkai and Zhang, Jiahui and Yao, Yao and Li, Shiwei and Fang, Tian and Quan, Long},
  booktitle={Proc. IEEE/CVF Conf. Comput. Vis. Pattern Recognit. (CVPR)},
  title={{ASLFeat}: Learning Local Features of Accurate Shape and Localization},
  year={2020},
  pages={6588-6597}}

@INPROCEEDINGS{wave-mlp,
  author={Tang, Yehui and Han, Kai and Guo, Jianyuan and Xu, Chang and Li, Yanxi and Xu, Chao and Wang, Yunhe},
  booktitle={Proc. IEEE/CVF Conf. Comput. Vis. Pattern Recognit. (CVPR)},
  title={An Image Patch is a Wave: Phase-Aware Vision {MLP}},
  year={2022},
  pages={10925-10934}}

@inproceedings{transformer,
  author    = {Vaswani, A. and Shazeer, N. and Parmar, N. and Uszkoreit, J. and Jones, L. and Gomez, A. N and Kaiser, L. and Polosukhin, I.},
  booktitle = {Proc. Adv. Neural Inf. Process. Syst. (NeurIPS)},
  pages     = {5998-6008},
  title     = {Attention is All you Need},
  year      = {2017}}

@ARTICLE{Changwei23,
  author={Wang, Changwei and Xu, Rongtao and Lu, Ke and Xu, Shibiao and Meng, Weiliang and Zhang, Yuyang and Fan, Bin and Zhang, Xiaopeng},
  journal={IEEE Trans. Pattern Anal. Mach. Intell.},
  title={Attention Weighted Local Descriptors},
  year={2023},
  volume={45},
  number={9},
  pages={10632-10649}}

@INPROCEEDINGS{ContextDesc,
  author={Luo, Zixin and Shen, Tianwei and Zhou, Lei and Zhang, Jiahui and Yao, Yao and Li, Shiwei and Fang, Tian and Quan, Long},
  booktitle={Proc. IEEE/CVF Conf. Comput. Vis. Pattern Recognit. (CVPR)},
  title={{ContextDesc}: Local Descriptor Augmentation With Cross-Modality Context},
  year={2019},
  pages={2522-2531}}

@INPROCEEDINGS{Oxford100k,
  author={Philbin, James and Chum, Ondrej and Isard, Michael and Sivic, Josef and Zisserman, Andrew},
  booktitle={Proc. IEEE/CVF Conf. Comput. Vis. Pattern Recognit. (CVPR)},
  title={Object retrieval with large vocabularies and fast spatial matching},
  year={2007},
  pages={1--8}}

@article{adam,
  title={Adam: A method for stochastic optimization},
  author={Kingma, Diederik P and Ba, Jimmy},
  journal={arXiv preprint arXiv:1412.6980},
  year={2014}
}

@INPROCEEDINGS{Truong2021,
  author={Truong, Prune and Danelljan, Martin and Van Gool, Luc and Timofte, Radu},
  booktitle={Proc. IEEE/CVF Conf. Comput. Vis. Pattern Recognit. (CVPR)},
  title={Learning Accurate Dense Correspondences and When to Trust Them},
  year={2021},
  pages={5710-5720}}

@ARTICLE{10148998,
  author={Ye, Zhichao and Bao, Chong and Zhou, Xin and Liu, Haomin and Bao, Hujun and Zhang, Guofeng},
  journal={IEEE Trans. Circuits Syst. Video Technol.},
  title={{EC-SfM}: Efficient Covisibility-Based Structure-From-Motion for Both Sequential and Unordered Images},
  year={2024},
  volume={34},
  number={1},
  pages={110-123},
}

@ARTICLE{10742393,
  author={Hu, Xinggang and Wu, Yanmin and Zhao, Mingyuan and Yang, Linghao and Zhang, Xiangkui and Ji, Xiangyang},
  journal={IEEE Trans. Circuits Syst. Video Technol.},
  title={{PAS-SLAM}: A Visual {SLAM} System for Planar-Ambiguous Scenes},
  year={2025},
  volume={35},
  number={3},
  pages={2026-2044},
}

@InProceedings{ASpanFormer,
author={Chen, Hongkai
and Luo, Zixin
and Zhou, Lei
and Tian, Yurun
and Zhen, Mingmin
and Fang, Tian
and McKinnon, David
and Tsin, Yanghai
and Quan, Long},
title={{ASpanFormer}: Detector-Free Image Matching with Adaptive Span {Transformer}},
booktitle = {Proc. Eur. Conf. Comput. Vis. (ECCV)},
year={2022},
pages={20--36}}

@INPROCEEDINGS{LightGlue,
  author={Lindenberger, Philipp and Sarlin, Paul-Edouard and Pollefeys, Marc},
  booktitle={Proc. Int. Conf. Comput. Vis. (ICCV)},
  title={{LightGlue}: Local Feature Matching at Light Speed},
  year={2023},
  pages={17581-17592},
}

@INPROCEEDINGS{ASTR,
  author={Yu, Jiahuan and Chang, Jiahao and He, Jianfeng and Zhang, Tianzhu and Yu, Jiyang and Wu, Feng},
  booktitle={Proc. IEEE/CVF Conf. Comput. Vis. Pattern Recognit. (CVPR)},
  title={Adaptive Spot-Guided {Transformer} for Consistent Local Feature Matching},
  year={2023},
  pages={21898-21908},
}

@inproceedings{scannet,
  title={{ScanNet}: Richly-annotated {3D} reconstructions of indoor scenes},
  author={Dai, Angela and Chang, Angel X and Savva, Manolis and Halber, Maciej and Funkhouser, Thomas and Nie{\ss}ner, Matthias},
  booktitle={Proc. IEEE Conf. Comput. Vis. Pattern Recognit. (CVPR)},
  pages={5828--5839},
  year={2017}
}

@inproceedings{AdaLAM,
  title={Handcrafted outlier detection revisited},
  author={Cavalli, Luca and Larsson, Viktor and Oswald, Martin Ralf and Sattler, Torsten and Pollefeys, Marc},
  booktitle = {Proc. Eur. Conf. Comput. Vis. (ECCV)},
  pages={770--787},
  year={2020},
}

@inproceedings{DiffGlue,
  title={{DiffGlue}: Diffusion-aided image feature matching},
  author={Zhang, Shihua and Ma, Jiayi},
  booktitle={Proc. ACM Int. Conf. Multimedia},
  pages={8451--8460},
  year={2024}
}

@inproceedings{resmatch,
  title={{ResMatch}: Residual attention learning for feature matching},
  author={Deng, Yuxin and Zhang, Kaining and Zhang, Shihua and Li, Yansheng and Ma, Jiayi},
  booktitle = {Proc. AAAI Conf. Artif. Intell. (AAAI)},
  pages={1501--1509},
  year={2024}
}

@ARTICLE{MatchMamba,
  author={Wu, Yubin and Li, Xiaojie and Chen, Hao and Yang, Changcai and Wei, Lifang and Chen, Riqing},
  journal={IEEE Trans. Circuits Syst. Video Technol.},
  title={{MatchMamba}: Correspondence Pruning via Selective State Space Model},
  year={2025}}

@ARTICLE{CGR-Net,
  author={Yang, Changcai and Li, Xiaojie and Ma, Jiayi and Zhuang, Fengyuan and Wei, Lifang and Chen, Riqing and Chen, Guodong},
  journal={IEEE Trans. Circuits Syst. Video Technol.},
  title={{CGR-Net}: Consistency Guided {ResFormer} for Two-View Correspondence Learning},
  year={2024},
  volume={34},
  number={12},
  pages={12450-12465}}

@article{MGCNet,
title = {{MGCNet}: Multi-granularity consensus network for remote sensing image correspondence pruning},
journal = {ISPRS J. Photogramm. Remote Sens.},
volume = {219},
pages = {38-51},
year = {2025},
author = {Zhuang, Fengyuan and Liu, Yizhang and Li, Xiaojie and Zhou, Ji and Chen, Riqing and Wei, Lifang and Yang, Changcai and Ma, Jiayi}
}

@INPROCEEDINGS{AdaMatcher,
  author={Huang, Dihe and Chen, Ying and Liu, Yong and Liu, Jianlin and Xu, Shang and Wu, Wenlong and Ding, Yikang and Tang, Fan and Wang, Chengjie},
  booktitle={Proc. IEEE/CVF Conf. Comput. Vis. Pattern Recognit. (CVPR)},
  title={Adaptive Assignment for Geometry Aware Local Feature Matching},
  year={2023},
  pages={5425-5434},
}

@article{OAMatcher,
title = {{OAMatcher}: An overlapping areas-based network with label credibility for robust and accurate feature matching},
journal = {Pattern Recognit.},
volume = {147},
pages = {110094:1--110094:14},
year = {2024},
author = {Dai, Kun and Xie, Tao and Wang, Ke and Jiang, Zhiqiang and Li, Ruifeng and Zhao, Lijun}}

@article{DeepMatcher,
title = {{DeepMatcher}: A deep transformer-based network for robust and accurate local feature matching},
journal = {Expert Syst. Appl.},
volume = {237},
pages = {121361},
year = {2024},
author = {Xie, Tao and Dai, Kun and Wang, Ke and Li, Ruifeng and Zhao, Lijun}}

@ARTICLE{VD-Matcher,
  author={Dai, Kun and Zhou, Zilong and Jiang, Zhiqiang and Sun, Qihao and Xie, Tao and Gao, Hongbo and An, Tao and Li, Ruifeng and Zhao, Lijun},
  journal={IEEE Trans. Circuits Syst. Video Technol.},
  title={{VD-Matcher}: A Very Deep Local Feature Matcher with Weight Recycling and Keypoint Detection},
  year={2025},
}

@ARTICLE{10794561,
  author={Li, Zizhuo and Ma, Jiayi},
  journal={IEEE Trans. Image Process.},
  title={Learning Feature Matching via Matchable Keypoint-Assisted Graph Neural Network},
  year={2025},
  volume={34},
  pages={154-169}}

@inproceedings{li2025comatch,
  title={{CoMatch}: Dynamic Covisibility-Aware Transformer for Bilateral Subpixel-Level Semi-Dense Image Matching},
  author={Li, Zizhuo and Lu, yifan and Tang, Linfeng and Zhang, Shihua and Ma, Jiayi},
  booktitle={Proc. Int. Conf. Comput. Vis. (ICCV)},
  year={2025},
  pages={18521--18530}}

@article{Survey2025,
      title={Deep Learning Reforms Image Matching: A Survey and Outlook},
      author={Zhang, Shihua and Li, Zizhuo and Zhang, Kaining and Lu, Yifan and Deng, Yuxin and Tang, Linfeng and Jiang, Xingyu and Ma, Jiayi},
      year={2025},
      eprint={2506.04619},
      journal={arXiv},
      url={https://arxiv.org/abs/2506.04619},
}

@article{oquab2023dinov2,
  title={{DINOv2}: Learning Robust Visual Features without Supervision},
  author={Oquab, Maxime and Darcet, Timoth¨¦e and Moutakanni, Theo and Vo, Huy V. and Szafraniec, Marc and Khalidov, Vasil and Fernandez, Pierre and Haziza, Daniel and Massa, Francisco and El-Nouby, Alaaeldin and Howes, Russell and Huang, Po-Yao and Xu, Hu and Sharma, Vasu and Li, Shang-Wen and Galuba, Wojciech and Rabbat, Mike and Assran, Mido and Ballas, Nicolas and Synnaeve, Gabriel and Misra, Ishan and Jegou, Herve and Mairal, Julien and Labatut, Patrick and Joulin, Armand and Bojanowski, Piotr},
  journal = {arXiv},
  url={https://arxiv.org/abs/2304.07193},
  year={2023}
}

@ARTICLE{U-Match,
  author={Li, Zizhuo and Zhang, Shihua and Ma, Jiayi},
  journal={IEEE Trans. Pattern Anal. Mach. Intell.},
  title={{U-Match}: Exploring Hierarchy-Aware Local Context for Two-View Correspondence Learning},
  year={2024},
  volume={46},
  number={12},
  pages={10960-10977},
}

@INPROCEEDINGS{MINIMA,
  author={Ren, Jiangwei and Jiang, Xingyu and Li, Zizhuo and Liang, Dingkang and Zhou, Xin and Bai, Xiang},
  booktitle={Proc. IEEE/CVF Conf. Comput. Vis. Pattern Recognit. (CVPR)},
  title={{MINIMA}: Modality Invariant Image Matching},
  year={2025},
  pages={23059-23068}}

%



\begin{IEEEbiography}[{\includegraphics[width=1in]{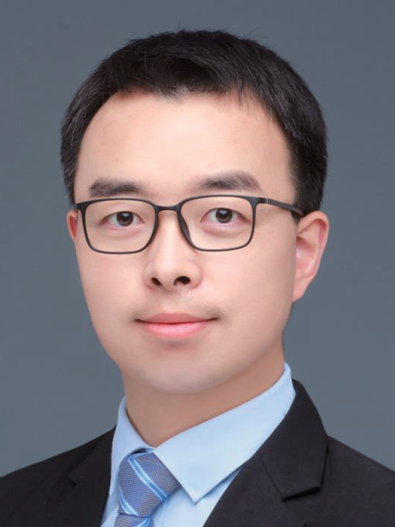}}]{Songlin Du}
received the Ph.D. degree in Physics from Lanzhou University, Lanzhou, China, and the second Ph.D. degree in Engineering from Waseda University, Tokyo, Japan, in 2019. He is currently an Associate Professor with the School of Automation, Southeast University, Nanjing, China. He was a recipient of the Best Paper Award from the IEEE International Symposium on Intelligent Signal Processing and Communication Systems (ISPACS) in 2017, the Best Presentation Award from the International Conference on Intelligent Systems and Image Processing (ICISIP) in 2018, the Excellent Paper Award from the International Collaboration Symposium on Information, Production and System (ISIPS) in 2018, and the Best Presentation Award from the Computer Vision Session of the International Conference on Artificial Intelligence and Virtual Reality (AIVR) in 2019. He is an author of the book \emph{Human Pose Analysis: Deep Learning Meets Human Kinematics in Video} (Singapore: Springer Nature, 2024). His research interests include computer vision and its applications.
\end{IEEEbiography}

\vfill

\begin{IEEEbiography}[{\includegraphics[width=1in]{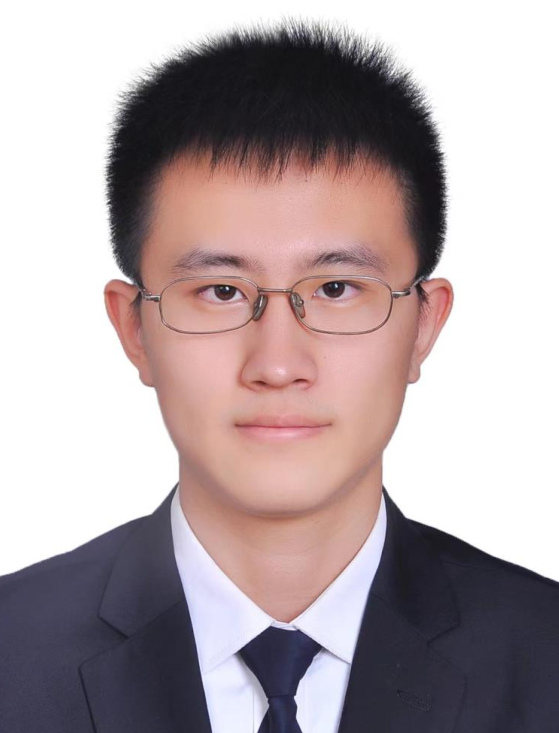}}]{Xiaoyong Lu}
received the B.E. degree in automation from Nanjing University of Science and Technology, Nanjing, China, in 2021. He is currently pursuing the doctoral degree in control science and engineering at Southeast University, Nanjing, China. His research interests include computer vision and machine learning.
\end{IEEEbiography}

\vfill

\begin{IEEEbiography}[{\includegraphics[width=1in]{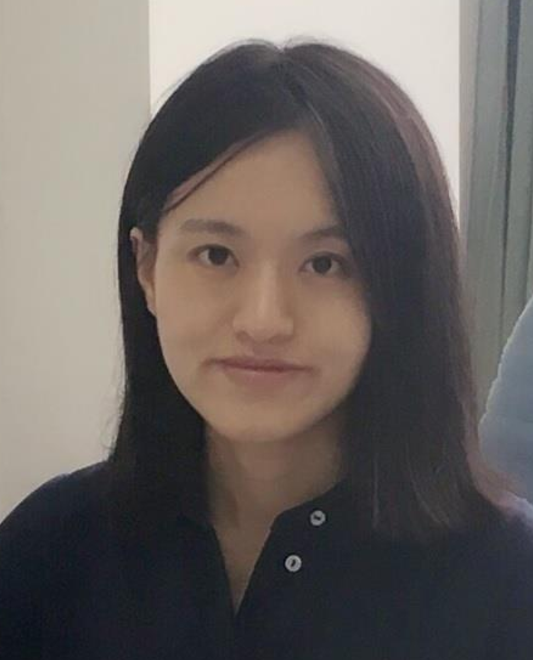}}]{Yaping Yan}
received the B.E. degree from Zhejiang University, Hangzhou, China, in 2010, the M.E. degree from Lanzhou University, Lanzhou, China, in 2016, and the Ph.D. degree from Hokkaido University, Sapporo, Japan, in 2020. She is currently with the School of Computer Science and Engineering, Southeast University, Nanjing, China. Her research focuses on computer vision and machine learning.
\end{IEEEbiography}

\vfill

\begin{IEEEbiography}[{\includegraphics[width=1in]{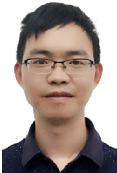}}]{Guobao Xiao}(Senior Member, IEEE)
received the Ph.D. degree in Computer Science and Technology from Xiamen University, Xiamen, China, in 2016. He is currently a Professor at Tongji University, Shanghai, China. He has published over 50 papers in journals and conferences including IEEE TPAMI/TIP, IJCV, ICCV, and ECCV. He was awarded the Best Ph.D. Thesis Award by China Society of Image and Graphics (a total of ten awardees in China). He also served on the program committee (PC) of CVPR, ICCV, ECCV, etc. His research interests include machine learning, computer vision and pattern recognition.
\end{IEEEbiography}

\vfill
\vfill

\begin{IEEEbiography}[{\includegraphics[width=1in]{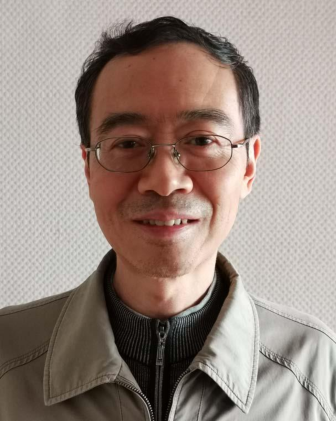}}]{Xiaobo Lu}
received the B.S. degree from Shanghai Jiao Tong University, Shanghai, China, the M.S. degree from Southeast University, Nanjing, China, and the Ph.D. degree from Nanjing University of Aeronautics and Astronautics, Nanjing, China. He did his postdoctoral research with the Chien-Shiung Wu Laboratory, Southeast University, from 1998 to 2000. He is currently a Professor with the School of Automation and the Deputy Director of the Detection Technology and Automation Research Institute, Southeast University. He is a co-author of the book \emph{An Introduction to the Intelligent Transportation Systems} (China Communications Press, Beijing, 2008). His research interests include image processing, signal processing, pattern recognition, and computer vision. He was the recipient of the first prize of the Natural Science Award from the Ministry of Education of China and the second prize of the Jiangsu Provincial Science and Technology Award.
\end{IEEEbiography}

\vfill

\begin{IEEEbiography}[{\includegraphics[width=1in]{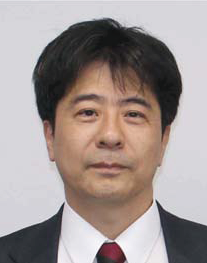}}]{Takeshi Ikenaga}(Senior Member, IEEE)
received his B.E. and M.E. degrees in electrical engineering and Ph.D degree in information computer science from Waseda University, Tokyo, Japan, in 1988, 1990, and 2002, respectively. He joined LSI Laboratories, Nippon Telegraph and Telephone Corporation (NTT) in 1990, where he had been undertaking research on the design and test methodologies for high performance ASICs, a real-time MPEG2 encoder chip set, and a highly parallel LSI system design for image understanding processing. He is presently a professor in the system integration field of the Graduate School of Information, Production and Systems, Waseda University. His current interests are image and video processing systems, which covers video compression (e.g. H.265/HEVC, SHVC, SCC), video filter (e.g. super resolution, noise reduction, high-dynamic range imaging), and video recognition (e.g. sports analysis, feature point detection, object tracking) and video communication (e.g. UWB, LDPC, public key encryption). He is a senior member of the Institute of Electrical and Electronics Engineers (IEEE), a member of the Institute of Electronics, Information and Communication Engineers of Japan (IEICE) and the Information Processing Society of Japan (IPSJ).
\end{IEEEbiography}





\end{document}